\documentclass[letterpaper]{article} 
\usepackage{aaai2026}  
\usepackage{times}  
\usepackage{helvet}  
\usepackage{courier}  
\usepackage[hyphens]{url}  
\usepackage{graphicx} 
\usepackage{natbib}  
\usepackage{caption} 
\usepackage{algorithm}
\usepackage{algorithmic}

\usepackage{lipsum}
\usepackage{amsthm}

\usepackage[inkscapelatex=false]{svg}
\usepackage{subfigure}
\usepackage{multirow}
\usepackage{placeins}
\usepackage{siunitx}
\usepackage{amsmath}
\usepackage{bm}
\usepackage{xcolor}
\usepackage{amssymb}
\usepackage{booktabs}
\usepackage{xspace}

\newcommand\zty[1]{{\color{black}#1}}
\newcommand\gqh[1]{{\color{black}#1}}

\newtheoremstyle{my_definition}
  {\topsep} 
  {\topsep} 
  {\itshape} 
  {0pt} 
  {\bfseries\itshape} 
  {\textmd{.}} 
  { } 
  {\textup{\thmname{#1}\thmnumber{ #2}:} \thmnote{#3}} 

\theoremstyle{my_definition}
\newtheorem{definition}{Definition}

\newcommand{\name}[0]{DST\xspace}

\usepackage{newfloat}
\usepackage{listings}
\DeclareCaptionStyle{ruled}{labelfont=normalfont,labelsep=colon,strut=off} 
\floatstyle{ruled}
\newfloat{listing}{tb}{lst}{}
\floatname{listing}{Listing}
\title{Dual-branch Spatial-Temporal Self-supervised Representation for \\ Enhanced Road Network Learning}
\author{
    Qinghong Guo\textsuperscript{\rm 1},
    Yu Wang\textsuperscript{\rm 1},
    Ji Cao\textsuperscript{\rm 1},
    Tongya Zheng\textsuperscript{\rm 1,2,3}\thanks{Corresponding author.},
    Junshu Dai\textsuperscript{\rm 1},\\
    Bingde Hu\textsuperscript{\rm 1,4},
    Shunyu Liu\textsuperscript{\rm 5},
    Canghong Jin\textsuperscript{\rm 2}
}
\affiliations{
    \textsuperscript{\rm 1}State Key Laboratory of Blockchain and Data Security, Zhejiang University\\
    \textsuperscript{\rm 2}Zhejiang Provincial Engineering Research Center for Real-Time SmartTech in Urban Security Governance, School of Computer and Computing Science, Hangzhou City University\\
    \textsuperscript{\rm 3}Hangzhou High-Tech Zone (Binjiang) Institute of Blockchain and Data Security\\
    \textsuperscript{\rm 4}Bangsun Technology\\
    \textsuperscript{\rm 5}Nanyang Technological Univerisity\\

    \{q.h\_guo, yu.wang, caoj25, djs, tonyhu\}@zju.edu.cn,\\
    doujiang\_zheng@163.com, shunyu.liu@ntu.edu.sg, jinch@hzcu.edu.cn
}

\usepackage{bibentry}

\begin{document}

\maketitle

\begin{abstract}
Road network representation learning (RNRL) has attracted increasing attention from both researchers and practitioners as various spatiotemporal tasks are emerging.
\zty{Recent advanced methods leverage Graph Neural Networks (GNNs) and contrastive learning to characterize the spatial structure of road segments in a self-supervised paradigm.
However, spatial heterogeneity and temporal dynamics of road networks raise severe challenges to the neighborhood smoothing mechanism of self-supervised GNNs.
To address these issues, we propose a \textbf{D}ual-branch \textbf{S}patial-\textbf{T}emporal self-supervised representation framework for enhanced road representations, termed as \name.
On one hand, \name designs a mix-hop transition matrix for graph convolution to incorporate dynamic relations of roads from trajectories.
Besides, \name contrasts road representations of the vanilla road network against that of the hypergraph in a spatial self-supervised way. The hypergraph is newly built based on three types of hyperedges to capture long-range relations.
On the other hand, \name performs next token prediction as the temporal self-supervised task on the sequences of traffic dynamics based on a causal Transformer, which is further regularized by differentiating traffic modes of weekdays from those of weekends.
}
Extensive experiments against state-of-the-art methods verify the superiority of our proposed framework. Moreover, the comprehensive spatiotemporal modeling facilitates \name to excel in zero-shot learning scenarios.
\end{abstract}

\begin{links}
    \link{Code}{https://github.com/chaser-gua/DST}
\end{links}

\section{Introduction}\label{sec:intro}
The road network is a vital infrastructure in a city, reflecting connectivity and accessibility between road segments, which guide human mobility. With the development of smart traffic systems~\cite{azgomi2018brief, camero2019smart,wang2024cola, wang2024STAR}, the road network has become a core component supporting various smart city applications, such as traffic inference~\cite{yang2022robust, zou2023novel} and destination prediction~\cite{xue2013destination,zong2019trip}. Road network representation learning (RNRL) aims to learn universal and low-dimensional vector representations for road networks to enhance performance on downstream tasks~\cite{NEURIPS2024_15cc8e4a}. Consequently, RNRL has gained significant attention from researchers and is emerging as a powerful tool for spatial-temporal management.

Early efforts leverage the natural graph topology of the road network to learn task-specific representations based on Graph Neural Networks (GNNs)~\cite{jepsen2019graph,li2020spatial,gharaee2021graph}. While these methods demonstrate robust performance on their original tasks, their efficacy often diminishes when transferred to different downstream tasks.
\zty{
Recent research on RNRL methods is inspired by powerful self-supervised learning paradigms to explore various geospatial relationships between roads through reconstruction and contrastive learning tasks~\cite{wu2020learning, chen2021robust}.
Additionally, abundant trajectory data are utilized to characterize the dynamic relationships between roads, augmenting road representations with auxiliary information~\cite{mao2022jointly, schestakov2023road}.
Nonetheless, the neighborhood smoothing mechanism inherent in GNNs leads to suboptimal road representations due to the spatial-temporal dynamics of roads, which adversely affects applications in dynamic downstream tasks.
}
\begin{figure}[tp!]
    \centering
    \subfigure[Spatial heterogeneity.]
    {\label{fig:intro-a}
    \includegraphics[height=0.28\linewidth,width=0.4\linewidth]{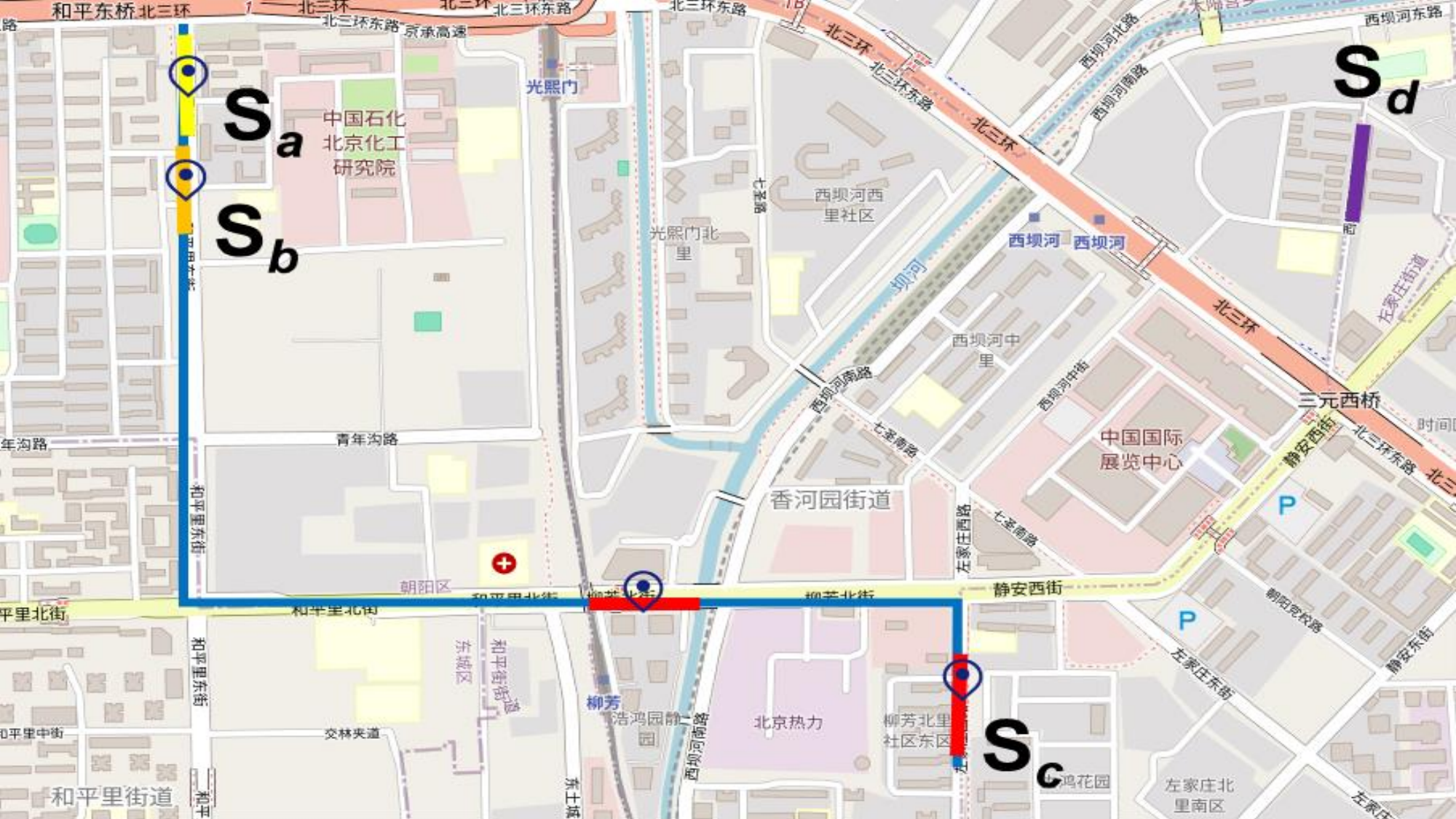}}
    \subfigure[Temporal dynamics.]{\label{fig:intro-b}
    \includegraphics[width=0.56\linewidth, height=0.28\linewidth]{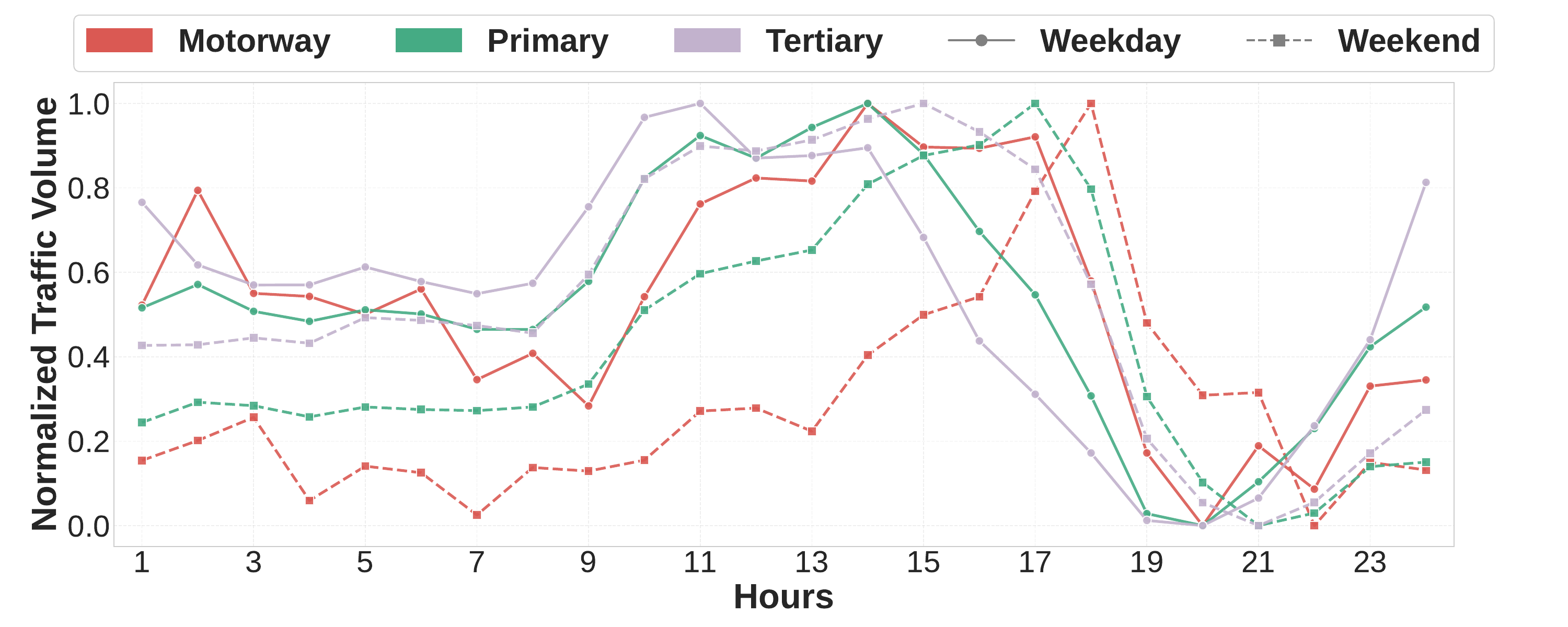}}
    \caption{\zty{An illustration example of spatial heterogeneity and temporal dynamics. (a) Distant roads with similar configurations can be connected by a travel trajectory, whereas nearby roads may not necessarily share similar characteristics. (b) The traffic patterns of roads are characterized not only by road types but also by temporal dynamics.}}
    \label{fig:intro}
\end{figure}

\zty{
Figure~\ref{fig:intro} illustrates the representation challenges of spatial heterogeneity and temporal dynamics for RNRL.}
\zty{Firstly, spatial heterogeneity reveals that road similarities can be estimated from several aspects beyond geospatial distance.
}
For example, as illustrated in Figure~\ref{fig:intro-a}, 
in the blue track, the next hop $S_b$ and $S_c$ after multiple hops from road $S_a$ have configurations similar to $S_a$. 
Extending the analysis to the broader road network shows that $S_d$, although spatially farther from $S_a$, serves a similar function (i.e., residential).
\zty{Hence, methods restricted by nearby roads may yield an incomplete understanding of distant roads.}
\zty{Secondly, temporal traffic dynamics at different hours serve as a crucial complement to road representations, which cannot be adequately captured by road networks alone.
}
As shown in Figure~\ref{fig:intro-b}, distinct road types exhibit markedly divergent traffic volumes temporally. Critically, dynamic patterns demonstrate pronounced weekday-weekend divergence even for identical road categories. 
\zty{Therefore, road representation necessitates the consideration of both spatial heterogeneity and temporal dynamics to effectively adapt to dynamic downstream tasks.}

\zty{To address these issues, we propose a \textbf{D}ual-branch \textbf{S}patial-\textbf{T}emporal self-supervised representation framework (\name) to obtain road representations from dual perspectives. 
From the spatial perspective, \name overcomes spatial heterogeneity by extracting dynamic relationships among roads using a mix-hop transition matrix, which is employed in the initial stage of graph convolution.
Additionally, we design a road hypergraph incorporating three types of hyperedges beyond the natural graph topology and integrate it with the standard road network through contrastive learning.
From the temporal perspective, \name utilizes next-token prediction for self-supervised representation based on a causal Transformer, along with a regularization task that differentiates traffic modes on weekdays from those on weekends.
\gqh{To accommodate heterogeneous inputs  while optimizing learning efficiency,}
the dual-branch representations are pre-trained separately and subsequently fused for downstream tasks.
}

Overall, our contributions can be summarized as below:
\begin{itemize}
    \item \zty{We propose a novel dual-branch road representation framework from both spatial and temporal views to address the challenges of spatial heterogeneity and temporal dynamics, called \name.}
    \item \zty{We design a mix-hop transition matrix and hypergraph-based contrastive learning for the spatial branch. Meanwhile, we devise next-token prediction and an auxiliary discrimination task for the temporal branch.}
    \item Extensive experiments on three datasets across three downstream tasks \zty{indicate the \emph{state-of-the-art} performance of \name based on superior road representation. Additionally, cross-city experiments demonstrate the strong transferability of our framework in zero-shot learning scenarios.}
\end{itemize}


\section{Preliminaries}\label{sec:pre}
In this section, we introduce the definitions of notation and descriptions of variables used in this paper.

\begin{definition}[Road Network]
     A road network can be represented as a directed graph $\mathcal{G} = \{\mathcal{R}, \mathcal E, X_\mathcal{R}, X_\mathcal E\}$, where $\mathcal{R}$ and $\mathcal{E}$ denote the set of road segments and their connections. For brevity, we let $N = \vert\mathcal{R}\vert$ denote the number of road segments and refer to the road segment as ``road'' in the later sections. $A_{\mathcal{G}} \in \mathbb R^{N \times N}$ is the adjacency matrix. If $\left(r_i, r_j\right)$ is reachable, then $A_{\mathcal{G}}\left[r_i, r_j\right] = 1$; otherwise $A_{\mathcal{G}}\left[r_i, r_j\right] = 0$. $X_\mathcal{R} \in \mathbb R^{N \times C_1}$ and $X_{\mathcal E} \in \mathbb{R}^{\vert \mathcal E \vert \times C_2}$ are the features of $\mathcal{R}$ retrieved from OpenStreetMap
     (OSM) and $\mathcal{E}$ computed from $X_\mathcal{R}$, where $C_1$ and $C_2$ represent the number of features.
\end{definition}

\begin{definition}[Road Hypergraph]
     A road hypergraph $\mathcal{G_H} = \left(\mathcal{R}, \mathcal{E_H}\right)$ consists of roads and hyperedges, where $\mathcal{E_H} = \left(e_1, e_2, \ldots, e_K\right)$ is the set of hyperedges. A simple graph is inadequate for capturing high-order relationships and functional properties between roads. Therefore, we introduce road hypergraph and utilize hyperedges to achieve these objectives. $A_{\mathcal{H}} \in \mathbb{R}^{N \times K}$ is the incidence matrix of a hypergraph. For $r_i$ and $e_k$, if $r_i \in e_k$, then $A_{\mathcal{H}}\left[r_i, e_k\right] = 1$; otherwise $A_{\mathcal{H}}\left[r_i, e_k\right] = 0$. 
\end{definition}

\begin{definition}[Trajectory]
    A trajectory $\tau \in \mathcal{T}$ is a sequence of continuously reachable $r$ in $\mathcal{G}$, denoted as $\tau = (r_1, r_2 \ldots, r_M)$. Trajectories capture movement patterns and contain rich dynamic information. In the following sections, $hop$ distance represents the number of hops within the trajectory, with $hop\left[r_i, r_j\right] = j-i$.
\end{definition}

\begin{definition}[Traffic Dynamics]
    The traffic dynamics $\mathcal{D_R} = \left[u_i\right]_{i=1}^{T} \in \mathbb{R}^{N \times T \times C}$ is the travel mode of $\mathcal{R}$ extracted from the trajectory. Let $u_i$ represent the number of visits on the road at the $i$-th time. The travel mode is divided by weekdays and weekends, with 24 hours in a day. Thus, we have $T = 24$ and $C = 2$. 
\end{definition}

\textbf{Problem Statement:} 
Given a road network $\mathcal{G}$, the goal of  Road Network Representation Learning is to learn a low-dimensional representation $v_r \in \mathbb{R}^{d}$ for each road, where $d$ represents the vector dimension. The learned representation can be utilized for various downstream tasks based on roads and trajectories.

\begin{figure*}
    \centering
    \includegraphics[width=\linewidth, height=0.45\linewidth]{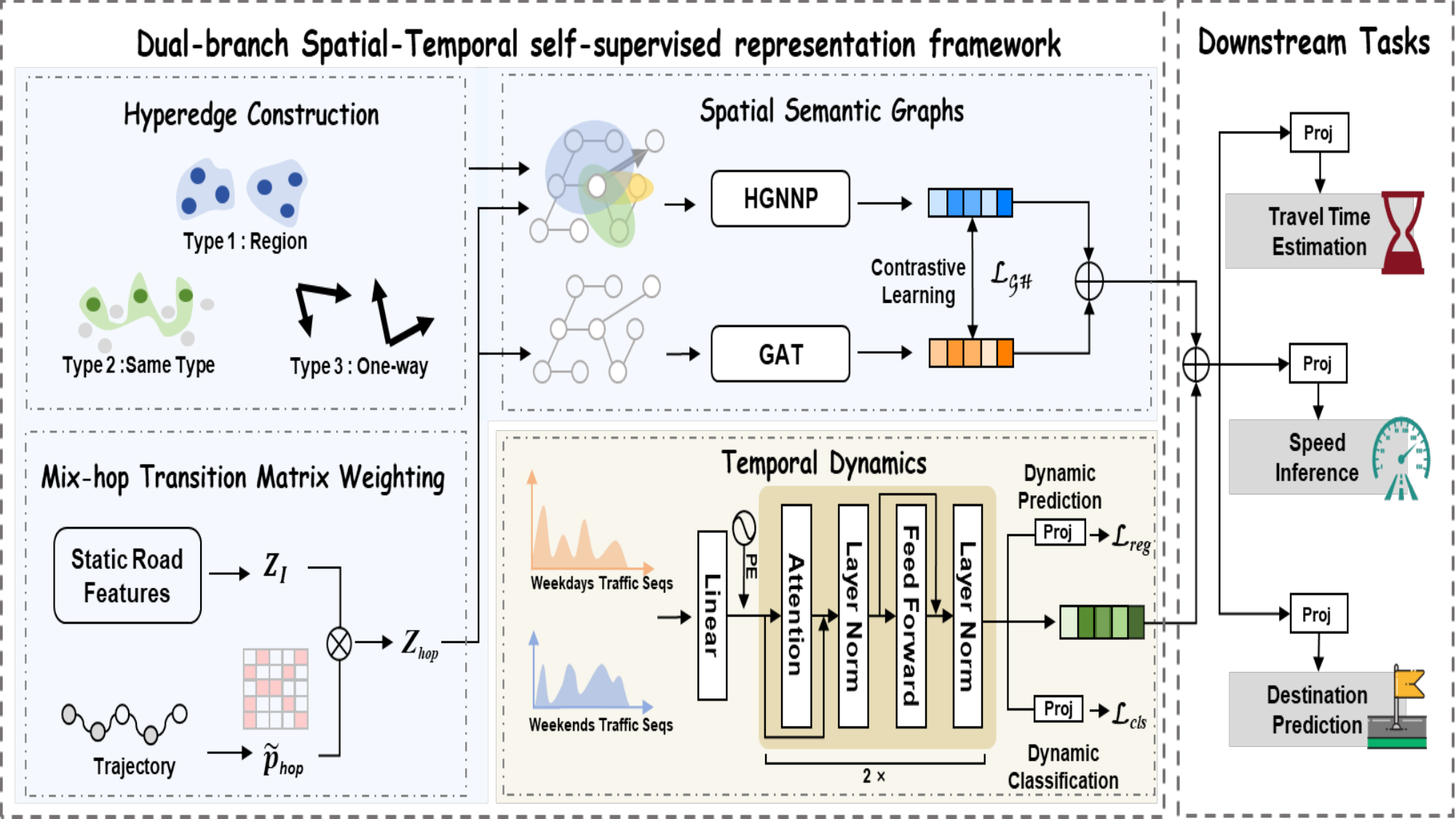}
    \caption{The overview of the proposed \name framework. The high-order relationships are modeled via mix-hop transition matrix weighting and multi-view graph contrastive learning. The temporal travel traffic dynamics are integrated by the Transformer with two specific task-driven updates. Both block co-enhanced representations power downstream tasks jointly.}
    \label{fig:frame}
\end{figure*}

\section{Methodology}\label{sec:metho}
In this section, we present the proposed \name framework. 
\gqh{\name overcomes spatial heterogeneity and captures high-order relationships by the learnable mix-hop transition matrix and hypergraph. 
Meanwhile, \name employs a Transformer-based traffic dynamic encoder and two self-supervised tasks to model the temporal travel traffic dynamics information of the roads. 
The overall architecture of the framework is shown in Figure~\ref{fig:frame}. }

\subsection{Mix-hop Transition Matrix Weighting}
Before all commencement, each static feature of the road network is embedded through an independent embedding layer. The initial embedding of a road segment is formed by concatenating the outputs of these layers:
\begin{equation}
    Z_{\mathcal{I}} = \Big\Vert_{c=1}^{C_1} \text{Emb}\left(X_{\mathcal{R}_c} \vert X_{\mathcal{R}_c} \in X_\mathcal{R}\right).
\end{equation}
Here, $\vert\vert$ is the vector concatenation operation and $Z_{\mathcal{I}}\in \mathbb{R}^{N \times d}$ is the feature embedding of all roads.

Next, we consider extracting the mix-hop transition matrix from trajectories on road networks. Unlike existing methods, our approach emphasizes the reachable and functional information of both the next hop and multiple hops of the road within the trajectory: 
\begin{equation}
    P_{hop}\left[r_i, r_j\right] = \sum_{\tau \in \mathcal{T}} \sum_{1\leq i<j\leq m} m-\left(j- i\right).
\end{equation}
Here, $m$ denotes the number of roads within the single trajectory. 
The weights are assigned based on hop distance.
In this context, the smaller hop distance receives larger initial weights, and the longer hop distance receives smaller initial weights. This strategy emphasizes adjacent links while incorporating reachable distant links.

To mitigate the impact of magnitude differences, we apply row normalization to the resulting transition matrix:
\begin{equation}
    \widetilde{P}_{hop}\left[r_i, r_j\right] = \frac{P_{hop}\left[r_i,r_j\right]}{\sum_{j=1}^{N}P_{hop}\left[r_i,r_j\right]}.
\end{equation}
Here, $\widetilde{P}_{hop} \in \mathbb{R}^{N \times N}$ is the final mix-hop transition matrix. If $\sum_{j=1}^N P_{hop}\left[r_i, r_j\right] = 0$, we set $\widetilde{P}_{hop}\left[r_i, r_i\right] = 1$. Then, we use it to initialize a learnable weight matrix, which is then used to update the representation further:
\begin{equation}
    Z_{hop} = \widetilde{P}_{hop} \cdot Z_\mathcal{I}. 
\end{equation}
Here, $\widetilde{P}_{hop}$ is the learnable mix-hop transition matrix and $Z_{hop} \in \mathbb{R}^{N\times d}$ is mix-hop weighted feature vectors.

\subsection{Spatial Semantic Graph Learning} 
Road networks exhibit heterogeneous characteristics, and the overemphasis of simple graph neural networks on topology may induce the over-smoothing problem\cite{chen2021robust, chen2024semantic}.
To mitigate this, we employ the spatial semantic hypergraph to capture high-order relationships between roads. Inspired by contrastive learning, we model road networks through graph-hypergraph views, updating parameters by maximizing mutual information (MI) between positive node pairs across representations.

\subsubsection{Spatial Graph View}
For the graph view, we employ a multi-layer Graph Attention Network (GAT)~\cite{velivckovic2017graph} to obtain the representation that captures the spatial topology of the road network.
At this stage, we also take into account some characteristics of road network connectivity $X_{\mathcal{E}}$. The specific formula for GAT is:
\begin{equation}
    Z_\mathcal{G} 
    =\Big\Vert_{h=1}^H \sigma\left(\sum_{j \in \mathcal{N}(i)} \alpha_{ij}^h \cdot W^hv_j\right),
\end{equation}

\begin{equation}
    \alpha_{ij} = \frac{\exp(\sigma(a^T \cdot \left[Wv_i\, \| \, Wv_j \, \| \, X_{\mathcal{E}_{ij}}\right]))}{\sum_{k \in \mathcal{N}\left(i\right)} \exp(\sigma(a^T \cdot \left[Wv_i\, \| \, Wv_k \, \| \, X_{\mathcal{E}_{ik}}\right]))},
\end{equation}
where $H$ represents the number of heads in the multi-head attention mechanism, $\sigma$ is the activation function, $W^h$ is the learnable parameter of $h$-th head, and $v_j \in {Z_{hop}}$ is the representation vector of the road $r_j$. 
The graph encoder generates the spatial representation $Z_\mathcal{G}$ for the graph view.

\subsubsection{Semantic HyperGraph View}
For the hypergraph view, as illustrated in Figure~\ref{fig:frame}, our hypergraph framework models complex semantic relationships through three specialized hyperedge types. 
$\mathcal{E_H}_1$ represents the functional zone generated through spectral clustering~\cite{von2007tutorial} of roads. $\mathcal{E_H}_2$ captures long-range dependencies by grouping roads of identical types, regardless of geographical proximity~\cite{zhang2023road}. $\mathcal{E_H}_3$ operationalizes Tobler's First Law of Geography~\cite{association1927annals} through hyperedges that connect geographically adjacent unidirectional roads.

We employ a multi-layer General Hypergraph Neural Networks (HGNN$^+$)~\cite{gao2022hgnn+} to further capture the functional semantics of the road network. By constructing hyperedges to represent various types of potential high-order and long-range correlations, HGNN$^+$ employs an adaptive hyperedge fusion strategy to effectively combine these correlations. Specifically, the formula for HGNN$^+$ is:
\begin{equation}
    Z_{\mathcal{H}} =  D\left(\mathcal{R}\right) \cdot A_\mathcal{H} \cdot D\left(\mathcal{E_H}\right) \cdot {A_\mathcal{H}}^T \cdot Z_{{hop}},
\end{equation}
\begin{equation}
    \mathcal{E_H} = \mathcal{E_{H}}_1 \parallel \mathcal{E_{H}}_2 \parallel \mathcal{E_{H}}_3,
\end{equation}
where the diagonal matrix $D(\mathcal{R})$ represents the degree of the nodes, denoting the number of hyperedges associated with the nodes, while $D(\mathcal{E_H})$ represents the degree of the hyperedges, indicating the number of nodes contained in each hyperedge, $\mathcal{E_H}$ is the set of three types hyperedges. 
The hypergraph encoder generates the semantic representation matrix $Z_\mathcal{H}$ for the hypergraph view.

\subsubsection{Contrastive Learning Loss}
We pursue the generalized rather than specific representation for downstream tasks through self-supervised learning. Therefore, we treat the representations $Z_\mathcal{G}$ and $Z_{\mathcal{H}}$ as two views of the road network for contrastive learning. We optimize the following MI maximization objective across these views~\cite{Zhu:2021tu}.
\begin{equation}
    \mathcal{L_{GH}} = -\frac{1}{N} \sum_{r_i \in \mathcal{R}} \left[ \frac{1}{|\mathcal{H}\left(r_i\right)|} \sum_{r_j \in \mathcal{H}\left(r_i\right)} I\left(v_{r_i}, h_{r_j}\right) \right],
\end{equation}
where $\mathcal{H}(r_i)$ is the set of nodes corresponding to the hypergraph view of $r_i$, $v_{r_i} \in Z_{\mathcal{G}}$ and $h_{r_j} \in Z_{\mathcal{H}}$ is the representation vector of $r_i$ and $r_j$ separately, and $I (\cdot)$ is the MI estimator~\cite{mao2022jointly}. The mini-batch strategy~\cite{li2014efficient} is implemented to optimize memory utilization.

\subsection{Temporal Dynamics Learning}
The coarse-grained road utilization feature fails to integrate into the temporal traffic flow characteristics. To this end, we employ a Transformer-based encoder to model fine-grained traffic dynamics. The joint optimization of the dynamic prediction and classification tasks enables comprehensive learning of 24-hour traffic evolution trends and patterns.

\subsubsection{Temporal Travel Dynamics Encoder}
The Transformer-based encoder ingests the previously defined traffic dynamic sequence, with its final hidden state serving as the compressed sequence representation. This architecture proves particularly effective for dynamic sequence modeling due to its self-attention mechanism, which captures both local fluctuations and global temporal dependencies:
\begin{equation}
    Z_{\mathcal{D}} = \text{TransEnc}\left(\text{PosEnc}\left(\mathcal{D_R}\right)\right)\left[-1\right].
\end{equation}
Here, $\text{PosEnc(·)}$ is sinusoidal positional encoding and $\text{TransEnc(·)}$ denotes a standard implementation of Transformer Encoder~\cite{waswani2017attention}. 
The temporal travel dynamic encoder produces the temporal representation $Z_\mathcal{D}$.

\subsubsection{Two-task Joint Dynamic Loss}
We aim to capture the patterns and trends of temporal traffic dynamic sequences through two tasks. Intuitively, the dynamic prediction task leverages the history sequence to predict the value of the sequence at the next time step, thereby capturing the long-term trend of the dynamic sequence. The dynamic classification task requires the model to categorize the input sequence into two types: weekday traffic and weekend traffic to distinguish between different dynamic sequence patterns. The corresponding regression value and classification probability are obtained after the representation passes through the project head, and these form the loss for the dynamic sequence task:
\begin{equation}
    \mathcal{L}_{reg} = \frac{1}{N \times C} \sum_{i=1}^{N \times C} \| y_i - \hat{y}_i \|^2,
\end{equation}
\begin{equation}
    \mathcal{L}_{cls} = \frac{1}{N \times C} \sum_{i=1}^{N \times C} \sum_{c=1}^{C} -{y}_i(c) \log(\hat{{y}}_i(c)),
\end{equation}
\begin{equation}
    \mathcal L_d = \lambda_{reg} \cdot \mathcal{L}_{reg} + \lambda_{cls} \cdot \mathcal{L}_{cls}.
\end{equation}
Here, $\hat{y_i}$ is the predicted output from the fully connected layer and $y_i$ is the ground truth, $\lambda_{reg}$ and $\lambda_{cls}$ are the hyperparameters to balance the two tasks.

\subsection{Representation Fusion for Downstream Tasks}
Finally, the spatial semantic representation and temporal dynamic representation are fused to form the final joint representation:
\begin{equation}
    Z = \text{Fusion} \left( Z_\mathcal{G}, Z_\mathcal{H}, Z_\mathcal{D} \right),
\end{equation}
where $Z_{\mathcal{G}}$, $Z_{\mathcal{H}}$, and $Z_{\mathcal{D}}$ are the updated spatial and temporal representations derived from the graph, hypergraph, and sequence encoder. $Z \in \mathbb{R}^{N \times d}$ is the final representation matrix of roads. 
The fusion methods we explore comprise concatenation along the last dimension, direct summation, and integration of representations via a gating mechanism. In this work, the first is employed for representation fusion.

This fusion method retains all the information of different representations, which is sufficiently effective for downstream tasks.
By incorporating temporal travel traffic dynamics into the road network, the final road representation becomes more informative, enhancing its effectiveness and versatility. Furthermore, our approach captures semantic relationships within road networks, which are essential for downstream tasks. 
For more fusion experiments and further discussion of fusion, please refer to Appendix D.1.

\section{Experiments}\label{sec:exper}
In this section, we present and analyze the experimental results across three real-world datasets and downstream tasks. 
The experimental results validate the effectiveness of leveraging spatiotemporal information from both trajectories and road networks to improve the quality of road network representations.
Due to space limitations, additional experimental details and results are provided in Appendices C and D.

\subsection{Experimental Setup}

\subsubsection{Datasets}
We use datasets from three real-world cities: Beijing, Porto, and Xi'an. The initial trajectory data consisted of GPS points collected by vehicles. We employ a map-matching algorithm~\cite{yang2018fast} to map it to the road sequence trajectory defined in the previous section. Please refer to Appendix C.1 for more details.

\subsubsection{Downstream Tasks}
The effectiveness of road representations is validated on three traffic-related downstream tasks: road speed inference (SI), travel time estimation (TE), and trajectory destination prediction (DP). A more detailed description can be found in Appendix C.2.

\subsubsection{Evaluation Metrics}
Following prior works~\cite{mao2022jointly,schestakov2023road}, we employ established evaluation metrics for each task: mean absolute error (MAE) and root mean square error (RMSE) for the SI and TE; and accuracy at 1 (ACC@1) and mean reciprocal rank (MRR) for the DP. These metrics are standard for their respective domains.

\subsubsection{Baseline Methods}
We evaluate \name against a series of baselines, including both classic graph learning methods and self-supervised RNRL methods. The former includes Node2Vec~\cite{grover2016node2vec}, GCN~\cite{kipf2016semi}, GAE~\cite{velivckovic2017graph} and T-GCN~\cite{zhao2019t}, while the latter comprises  SRN2Vec~\cite{wang2020representation}, Toast~\cite{chen2021robust}, JCLRNT~\cite{mao2022jointly}, TrajRNE~\cite{schestakov2023road} and DyToast~\cite{chen2024semantic}. See Appendix C.3 for more details.

\begin{table*}[htp!]
\centering
\resizebox{2.1\columnwidth}{!}{
\begin{tabular}{lcccccccccccc}
\toprule
\multirow{3.5}{*}{Methods} & \multicolumn{4}{c|}{Beijing} & \multicolumn{4}{c|}{Porto} & \multicolumn{4}{c}{Xi'an}\\
\cmidrule(lr){2 - 13}
 & 
\multicolumn{2}{c|}{Destination Prediction} & 
\multicolumn{2}{c|}{Travel Time Estimation} & 
\multicolumn{2}{c|}{Destination Prediction} & 
\multicolumn{2}{c|}{Travel Time Estimation} & 
\multicolumn{2}{c|}{Destination Prediction} & 
\multicolumn{2}{c}{Travel Time Estimation}\\
\cmidrule(lr){2 - 13}
& ACC@1 ($\mathrel{\uparrow}$)& \multicolumn{1}{c|}{MRR ($\mathrel{\uparrow}$)} & MAE ($\mathrel{\downarrow}$) & \multicolumn{1}{c|}{RMSE ($\mathrel{\downarrow}$)} & ACC@1 ($\mathrel{\uparrow}$) & \multicolumn{1}{c|}{MRR ($\mathrel{\uparrow}$)} & MAE ($\mathrel{\downarrow}$) & \multicolumn{1}{c|}{RMSE ($\mathrel{\downarrow}$)} & ACC@1 ($\mathrel{\uparrow}$) & \multicolumn{1}{c|}{MRR ($\mathrel{\uparrow}$)} & MAE ($\mathrel{\downarrow}$) & {RMSE ($\mathrel{\downarrow}$)}\\
\midrule
Node2Vec & 0.1954 & 0.2884 & 253.0633 & \multicolumn{1}{c|}{378.8966} & 0.2201 & 0.3364 & 109.5741 & \multicolumn{1}{c|}{150.4529} & 0.4088 & 0.5083 & 287.3482 & 430.5943 \\
GCN & 0.2189 & 0.2995 & 256.2389 & \multicolumn{1}{c|}{381.1153} & 0.3780 & 0.4861 & 108.4090 & \multicolumn{1}{c|}{151.8520} & 0.4054 & 0.4904 & 249.5978 & 387.8792 \\
GAE & 0.2472 & 0.3254 & 249.8528 & \multicolumn{1}{c|}{376.1648} & 0.3997 & 0.5165 & 107.6947 & \multicolumn{1}{c|}{150.9522} & 0.4223 & 0.5091 & 283.0110 & 417.5603 \\
TGCN & 0.2080 & 0.2911 & 274.5425 & \multicolumn{1}{c|}{402.6056} & 0.4584 & 0.5827 & 111.8000 & \multicolumn{1}{c|}{155.3018} & 0.3887 & 0.4823 & 231.3772 & 341.2320 \\
SRN2Vec & 0.4158 & 0.4702 & 242.3774 & \multicolumn{1}{c|}{368.4888} & 0.6552 & 0.7745 & 101.9539 & \multicolumn{1}{c|}{\underline{144.1221}} & 0.6966 & 0.7503 & 236.8440 & 378.1186 \\
Toast & 0.3031 & 0.3652 & 248.4413 & \multicolumn{1}{c|}{374.0255} & 0.6142 & 0.7324 & 101.7503 & \multicolumn{1}{c|}{145.7847} & 0.7103 & 0.7673 & 244.5761 & 381.7282\\
JCLRNT & 0.4222 & 0.5528 & 246.7876 & \multicolumn{1}{c|}{373.1402} & 0.5133 & 0.6626 & \underline{101.6065} & \multicolumn{1}{c|}{145.2937} & 0.6752 & 0.7711 & 240.6833 & 380.1059 \\
TrajRNE & \underline{0.6728} & \underline{0.7603} & \underline{237.1361} & \multicolumn{1}{c|}{\underline{363.2190}} & \underline{0.6728} & \underline{0.8063} & 102.5061 & \multicolumn{1}{c|}{145.2530} & \underline{0.8260} & \underline{0.8831} & \underline{207.6418} & \underline{314.7823} \\
DyToast & 0.4440 & 0.5164 & 239.6679 & \multicolumn{1}{c|}{365.3374} & 0.4887 & 0.6025 & 102.3489 & \multicolumn{1}{c|}{145.3401} & 0.7508 & 0.8080 & 238.4887 & 379.9900  \\
Ours & $\textbf{0.7288}$ & $\textbf{0.8213}$ & $\textbf{236.6965}$ & \multicolumn{1}{c|}{$\textbf{363.2039}$} & $\textbf{0.6766}$ & $\textbf{0.8101}$ & $\textbf{101.4223}$ & \multicolumn{1}{c|}{$\textbf{143.6598}$} & $\textbf{0.8335}$ & $\textbf{0.8950}$ & $\textbf{202.8479}$ & $\textbf{307.7553}$ \\
\bottomrule
\end{tabular}
}
\caption{The results on three real-world datasets in terms of the destination prediction task and the travel time estimation task. The best one is denoted by \textbf{bold} and the second-best is denoted by \underline{underline}. $\mathrel{\uparrow}$ and $\mathrel{\downarrow}$ denote higher is better and lower is better.}
\label{tab:performance_comparison-road}
\end{table*}

\begin{table}[tp!]
\centering
\resizebox{\columnwidth}{!}{
\begin{tabular}{lcccccc}
\toprule
\multirow{2.5}{*}{Methods} & \multicolumn{2}{c|}{Beijing} & \multicolumn{2}{c|}{Porto} & \multicolumn{2}{c}{Xi'an}\\
\cmidrule(lr){2 - 7}
& MAE  ($\mathrel{\downarrow}$) & \multicolumn{1}{c|}{RMSE  ($\mathrel{\downarrow}$)} & MAE  ($\mathrel{\downarrow}$) & \multicolumn{1}{c|}{RMSE  ($\mathrel{\downarrow}$)} & MAE  ($\mathrel{\downarrow}$) & \multicolumn{1}{c}{RMSE  ($\mathrel{\downarrow}$)} \\
\midrule
Node2Vec & 8.2268 & \multicolumn{1}{c|}{8.8945} & 9.0367 & \multicolumn{1}{c|}{10.1075} & 6.8403 & 8.8547 \\
GCN & 8.1926 & \multicolumn{1}{c|}{8.8436} & 8.9723 & \multicolumn{1}{c|}{10.0485} & 6.8014 & 8.8107 \\
GAE & 8.1600 & \multicolumn{1}{c|}{8.8154} & 8.9442 & \multicolumn{1}{c|}{10.0804} & 6.8040 & 8.8114 \\
TGCN & 5.8026 & \multicolumn{1}{c|}{6.7065} & 6.5242 & \multicolumn{1}{c|}{7.8042} & 6.0006 & 8.0067 \\
SRN2Vec & 6.9959 & \multicolumn{1}{c|}{7.8806} & 7.0657 & \multicolumn{1}{c|}{8.3563} & 6.0201 & 7.9983 \\
Toast & 6.6538 & \multicolumn{1}{c|}{7.4272} & 7.2837 & \multicolumn{1}{c|}{8.4783} & 5.5102 & 7.5338 \\
JCLRNT & \underline{2.8512} & \multicolumn{1}{c|}{\underline{3.9013}} & \underline{3.7475} & \multicolumn{1}{c|}{\underline{4.9999}} & \underline{4.5138} & \underline{5.7294} \\
TrajRNE & 3.0756 & \multicolumn{1}{c|}{4.3049} & 4.7854 & \multicolumn{1}{c|}{6.3375} & 5.1898 & 7.1767 \\
DyToast & 8.0957 & \multicolumn{1}{c|}{8.8124} & 8.6818 & \multicolumn{1}{c|}{9.8597} & 6.7461 & 8.7050 \\
Ours & $\textbf{2.4595}$ & \multicolumn{1}{c|}{$\textbf{3.2557}$} & $\textbf{3.4259}$ & \multicolumn{1}{c|}{$\textbf{4.5538}$} & $\textbf{4.4987}$ & $\textbf{5.6557}$ \\
\bottomrule
\end{tabular}
}
\caption{The results on three real-world datasets in terms of the speed inference task. The best one is denoted by \textbf{bold} and the second-best is denoted by \underline{underline}.}
\label{tab:performance_comparison-traj}
\end{table}

\subsection{Overall Performance}
Table~\ref{tab:performance_comparison-road} and~\ref{tab:performance_comparison-traj} present the mean values of the results obtained from five random seeds for \name and baseline models across different downstream tasks. We can find that \name outperforms the baseline models in all tasks, 
demonstrating its effectiveness in modeling high-order and long-range information about road network function and movement, as well as its superiority in integrating road network dynamic travel traffic.
Among the baseline methods, Node2Vec, GCN, and GAE are not specifically designed for road networks. They are limited to modeling simple graph structures and do not capture the intricate relationships between roads, leading to suboptimal performance in tasks. This highlights the importance of unique designs tailored for road networks.
Although TGCN models temporal traffic dynamics, its architectural paradigm is not optimized specifically for road networks. Furthermore, it fails to capture the functional semantics of road networks.
SRN2Vec and Toast apply random walks for road weighting, thus achieving better performance than the traditional graph learning methods. 
DyToast augments Toast with trigonometric time features, enhancing performance in destination prediction and travel time estimation tasks. However, its degraded performance in the speed inference task reveals representational robustness limitations. Suboptimal dynamic modeling approaches may degrade performance.
In contrast, JCLRNT and TrajRNE leverage trajectory information to enhance road network representations, achieving better performance across tasks than other methods.
Nevertheless, they are deficient in modeling temporal dynamics, while \name mines the long-range dependencies of roads in trajectory and models both spatial and temporal traffic representations of the road network, achieving superior performance compared to other methods.

\subsection{Ablation Studies}
\name employs the following core designs for road network representation: the mix-hop pattern extraction strategy initializes the learnable transition matrix to capture road long-range dependencies contained in trajectories; the hypergraph with different hyperedges models higher-order functional information; the dynamic travel traffic integration supplements temporal information. 
To evaluate the impact of these designs, we conduct ablation studies by removing specific components. (1) \emph{\underline{w/o $P_{hop}$:}} this variant excludes the learnable transition matrix; 
(2) \emph{\underline{w/o ${hg}_1$:}} this variant eliminates region relationship hyperedges from the hypergraph structure; 
(3) \emph{\underline{w/o ${hg}_2$:}} this variant eliminates the same type relationship hyperedges from the hypergraph structure; 
(4) \emph{\underline{w/o ${hg}_3$:}} this variant eliminates one-way relationship hyperedges from the hypergraph structure; 
(5) \emph{\underline{w/o tm:}} this variant omits temporal travel traffic dynamic modeling.

\begin{figure}[tp!]
    \centering
    \includegraphics[width=\linewidth]{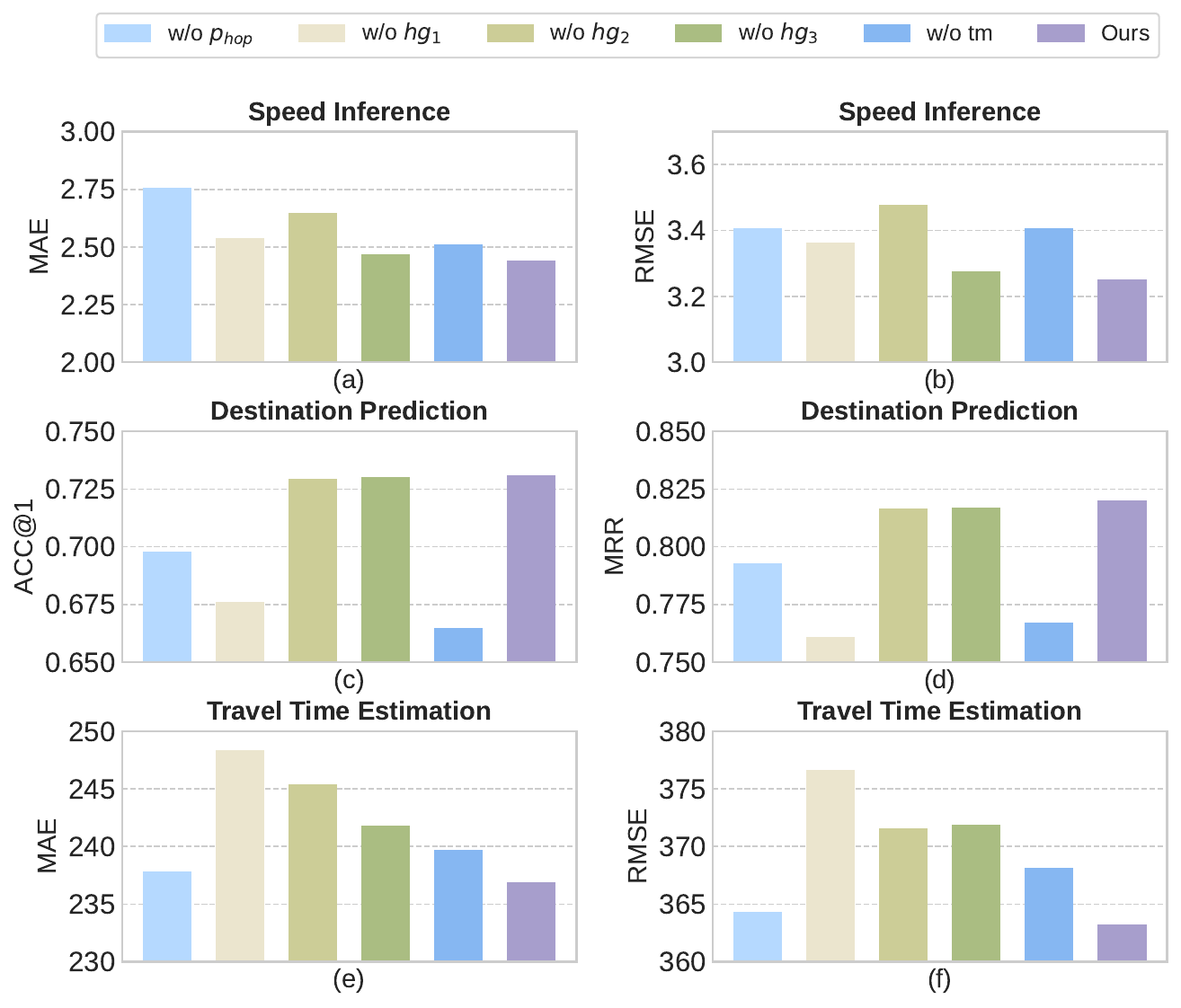}
    \caption{Ablation study on Beijing dataset.}
    \label{fig:ab-bj}
\end{figure}

Figure~\ref{fig:ab-bj} illustrates the performance of the model in the Beijing dataset after the removal of different design components. The results on other datasets are similar. It can be observed that removing any of these components results in inferior performance on downstream tasks compared to the complete \name. Specifically, we make the following analysis.
Firstly, the variants \emph{w/o $P_{hop}$} and \emph{w/o ${hg}_2$} exhibit the most severe performance degradation in the speed inference task, indicating that higher-order functional and motion relationships between roads are essential for learning effective road representations.
Secondly, the variant \emph{w/o tm} demonstrates significantly degraded performance across both trajectory-based tasks, underscoring the critical complementary role of dynamic travel traffic information in road network characterization.
Finally, the variants \emph{w/o ${hg}_1$} and \emph{w/o ${hg}_3$} exhibit moderate performance degradation across all tasks, indicating that diverse hyperedge types mutually reinforce each other to comprehensively model road network higher-order relationships. Please refer to Appendix D.2 for more details.


\subsection{Parameter Sensitivity}
\begin{figure}[tp!]
    \centering
    \includegraphics[width=\linewidth]{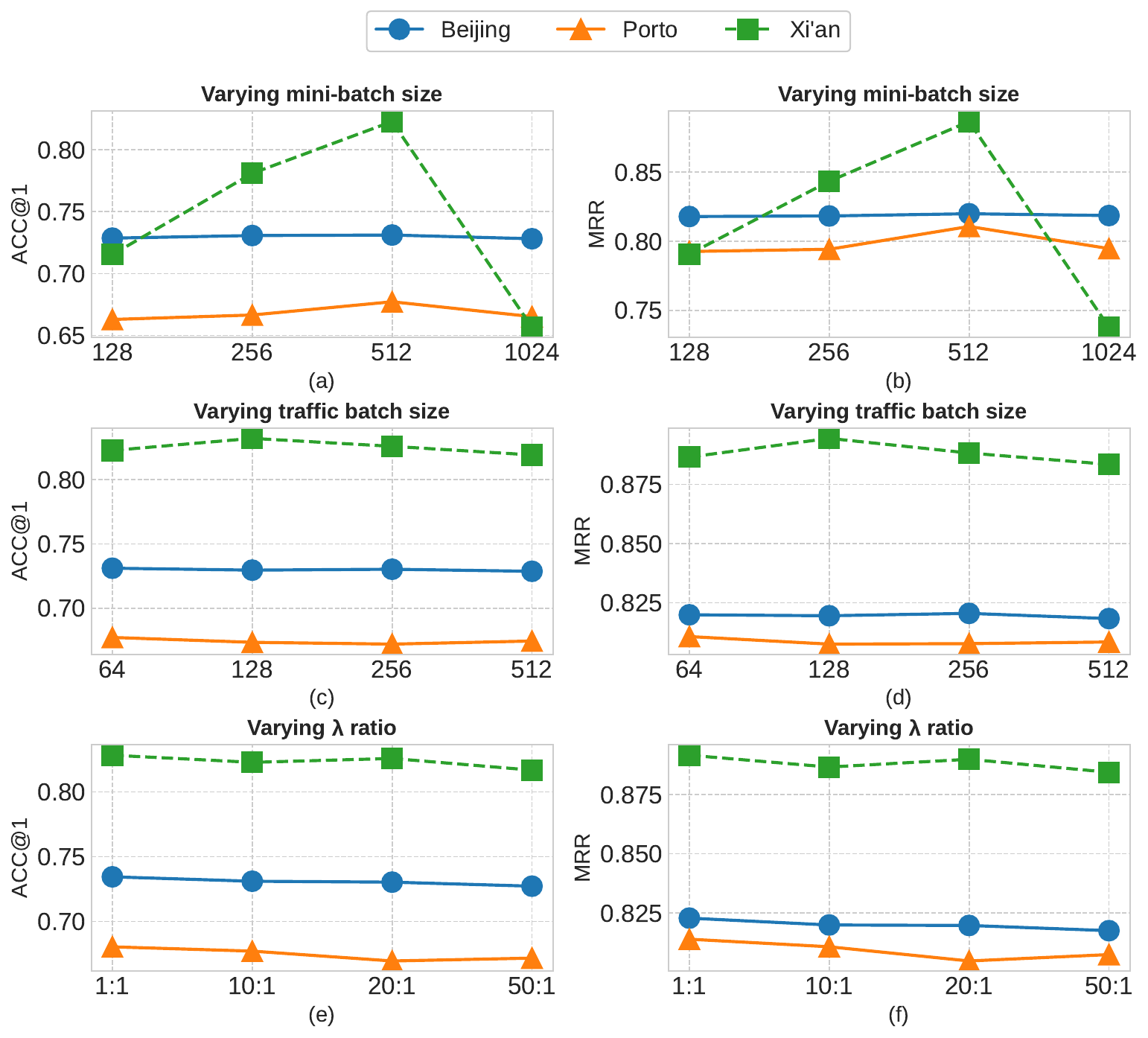}
    \caption{Parameter sensitivity on destination prediction.}
    \label{fig:minibatch}
\end{figure}

In this section, we analyze the effect of parameters on \name performance: the mini-batch size, the traffic batch size, and the ratio of $\lambda_{reg}$ and $\lambda_{cls}$. The evaluation is conducted on the destination prediction task for the three datasets. The results are shown in Figure~\ref{fig:minibatch}. We can find that \name maintains consistent excellent performance across all datasets despite parameter variations. The specific conclusions are as follows. Please refer to Appendix D.5 for more details.

Firstly, the results of \emph{\underline{varying mini-batch size}} indicate that a smaller size reduces the number of negative samples, potentially causing insufficient sample discrimination and underfitting. Conversely, a larger size increases computational complexity and memory demands while hindering effective learning. Thus, a moderate batch size achieves an optimal balance between learning efficiency and model performance. 
Secondly, the results of \emph{\underline{varying traffic batch size}} indicate that \name is not sensitive to this parameter. Nonetheless, there is a slight performance improvement in small-sized traffic batches, potentially due to the sparsity of the traffic sequence. A larger batch scale may introduce more noise.
Finally, the results of \emph{\underline{varying the $\lambda$ ratio}} indicate that increasing the next token prediction loss weight enhances model performance. This adjustment compensates for the significant magnitude difference in initial training losses between tasks, thereby improving task balance during training.

\subsection{Generalization on Zero-Shot Learning Scenarios}
Given substantial variations in urban systems, such as network topology, traffic patterns, and human mobility, transferable models significantly reduce deployment costs and resource expenditures across cities.
To evaluate the cross-city transferability of \name, we conduct a zero-shot learning experiment. Recognizing dependencies between the mix-hop matrix and urban nodes, we remove this component and train exclusively in Beijing. Evaluation then occurs on the Porto for three downstream tasks. In this context, we compare \name-transfer with GAE and JCLRNT, which support zero-shot learning. Table~\ref{tab:zero} demonstrates competitive zero-shot performance of \name attributable to comprehensive spatiotemporal representation learning.

\begin{table}[htp!]
\centering
\tiny
\setlength{\tabcolsep}{1mm}
\begin{tabular}{lcccccc}
\toprule
\multirow{2.5}{*}{Methods} & \multicolumn{2}{c|}{Speed Inference} & \multicolumn{2}{c|}{Destination Prediction} & \multicolumn{2}{c}{Travel Time Estimation}\\
\cmidrule(lr){2 - 7}
& MAE  ($\mathrel{\downarrow}$) & \multicolumn{1}{c|}{RMSE  ($\mathrel{\downarrow}$)} & ACC@1  ($\mathrel{\uparrow}$) & \multicolumn{1}{c|}{MRR  ($\mathrel{\uparrow}$)} & MAE  ($\mathrel{\downarrow}$) & \multicolumn{1}{c}{RMSE  ($\mathrel{\downarrow}$)} \\
\midrule
GAE & 9.0482 & \multicolumn{1}{c|}{10.1209} & 0.3119 & \multicolumn{1}{c|}{0.4204} & 112.1389 & 154.2606 \\
JCLRNT & 4.1047 & \multicolumn{1}{c|}{5.3195} & 0.0167 & \multicolumn{1}{c|}{0.0338} & 109.7691 & 151.6060 \\
\name-transfer & 3.5126 & \multicolumn{1}{c|}{4.6181} & 0.6424 & \multicolumn{1}{c|}{0.7765} & 108.0329 & 150.2018 \\
\bottomrule
\end{tabular}
\caption{Zero-shot learning results (Beijing → Porto).}
\label{tab:zero}
\end{table}

\subsection{Case Study}
\begin{figure}[tp!]
    \centering
    \subfigure[Distance to the anchor road.]
    {\label{fig:cs-a}
    \includegraphics[width=0.48\linewidth]{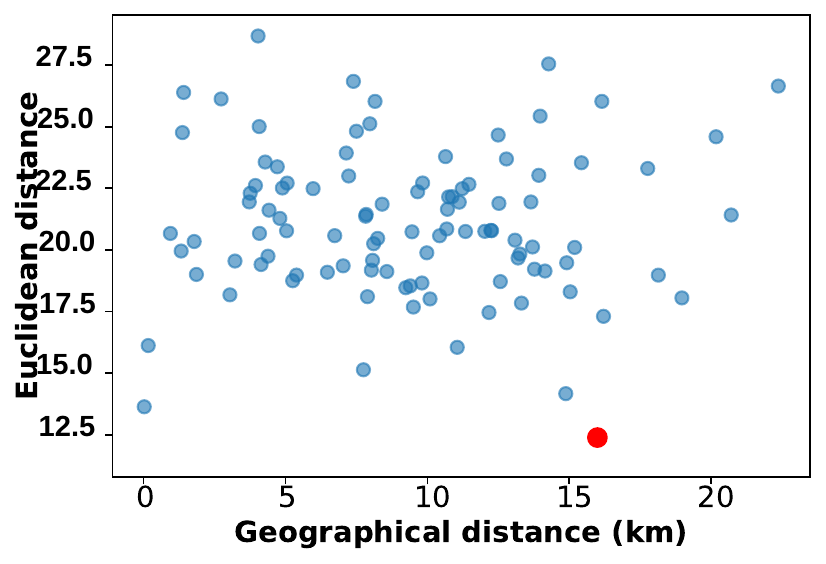}}
    \subfigure[Cases in the road network.]{\label{fig:cs-b}
    \includegraphics[width=0.44\linewidth, height=0.35\linewidth]{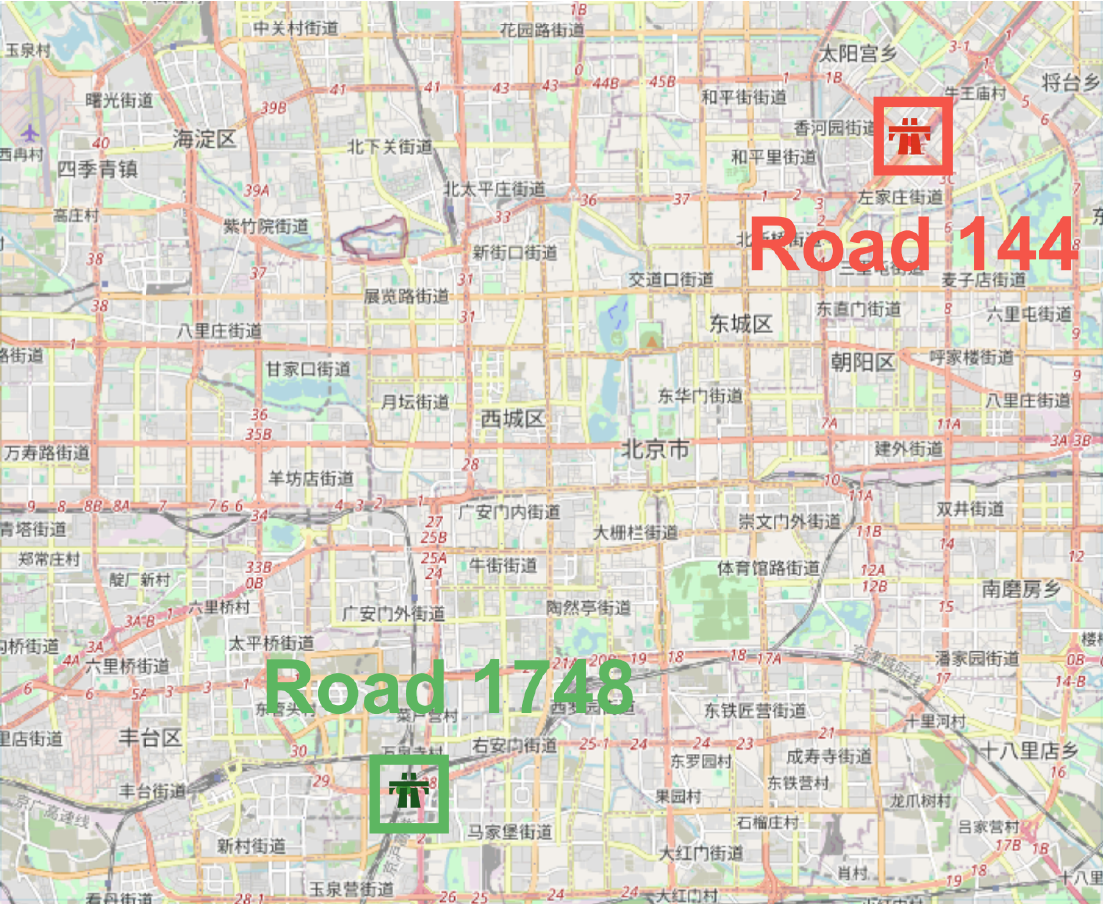}} \\
    \subfigure[The highest traffic road detailed visualization.]{\label{fig:cs-c-1}
    \includegraphics[width=0.95\linewidth, height=0.15\linewidth]{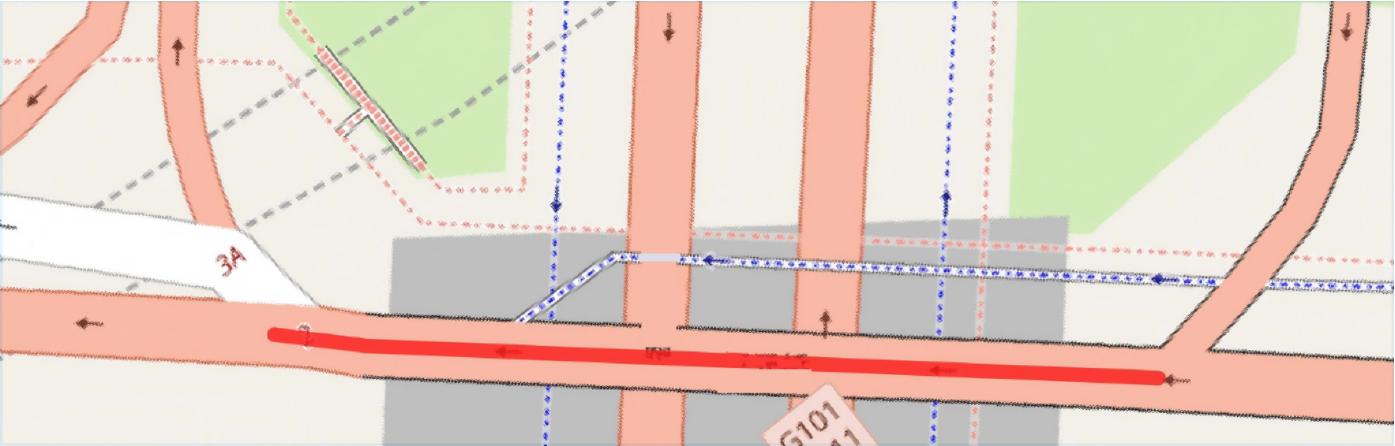}}
    \subfigure[The road geographically distant but representationally close.]{\label{fig:cs-c-2}
    \includegraphics[width=0.95\linewidth, height=0.15\linewidth]{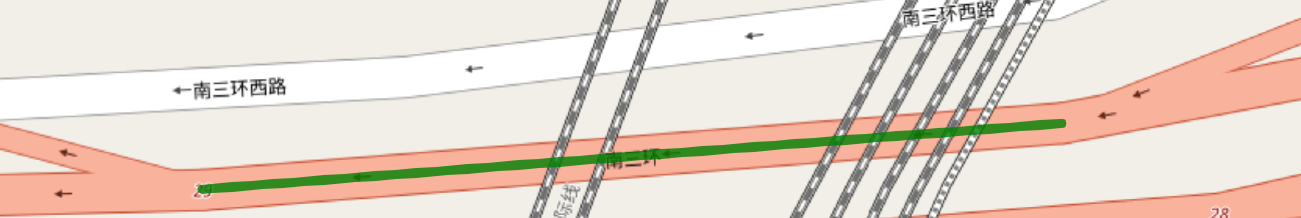}}
    \caption{Case study of the highest traffic road in Beijing.} 
    \label{fig:cs}
\end{figure}

In this section, we conduct a case study to demonstrate the ability of \name to capture high-order relationships between road segments. 
We choose the highest-traffic road (ID=144) in Beijing as an anchor road, and compute representation and geographic distances to other roads. Figure~\ref{fig:cs-a} shows the distance distribution of other roads and the anchor road. We identified a distant road segment (ID=1748) exhibiting close representational proximity. In Figure~\ref{fig:cs-b}, we visualize them in the road network. The anchor road is red, and the selection road is green. The visualization reveals that both the anchor and selected segments lie on the third ring of Beijing. Fine-grained analysis in Figure~\ref{fig:cs-c-1} and ~\ref{fig:cs-c-2} shows they share identical functional roles, unidirectional flow patterns, and dual inflow and outflow characteristics. This confirms \name successfully captures functional and traffic-dynamic relationships beyond spatial proximity.

\section{Related Work}\label{sec:rw}
\subsubsection{Road Network Representation Learning.}
Road networks are inherently structured as graphs. Most existing approaches draw inspiration from graph representation learning~\cite{zheng2022transition, zheng2023temporal,wang2024stega, yanggnns} and employ graph neural networks (GNNs) for modeling. Numerous GNN-based methods ~\cite{wu2020learning, pei2020geom, zhang2023road, cao2025holistic} develop specialized graph encoders to capture the spatial topology of road networks. 
As trajectory data proliferates, trajectory-enhanced methods ~\cite{mao2022jointly, schestakov2023road} extract movement and transfer patterns to advance road network representation learning.
However, these methods inadequately model higher-order information within functional and movement.
Our method integrates spatial semantic information from the perspectives of transfer mode and functional structure while incorporating travel traffic dynamics.

\subsubsection{Graph and Hypergraph Learning.}
Early graph learning~\cite{perozzi2014deepwalk, grover2016node2vec} focuses on the topology of graphs and learns the shallow representations of graphs. 
With the development of deep learning, graph convolutional network~\cite{kipf2016semi} and graph attention network~\cite{velivckovic2017graph} achieve significant success in graph learning by aggregating features in different ways. Temporal graph convolutional network considers dynamic changes in graphs~\cite{zhao2019t, wang2024unveiling} later. Limited by the adjacency in graphs, these models struggle to represent higher-order relationships between nodes. Therefore, hypergraph learning is employed to model higher-order relationships among two or more entities. Hypergraph Neural Networks~\cite{feng2019hypergraph,gao2022hgnn+} adopt different message-passing mechanisms to learn hypergraph representations. However, neither graphs nor hypergraphs can be directly applied to road networks that require tailored designs due to their complex spatiotemporal characteristics.

\subsubsection{Spatial-temporal Self-supervised Learning.}
Graph self-supervised learning methods are categorized into generative and contrastive paradigms. Generative approaches~\cite{bondy1977graph} leverage intrinsic graph structure as supervision, focusing on reconstruction tasks. Contrastive methods~\cite{you2020graph, liu2023CIA} operate on augmented graph views, exploiting invariance across views and discriminative information between negative pairs as supervisory signals.
Sequence self-supervised learning typically employs predictive pretext tasks, where supervisory signals are autonomously generated through statistical heuristics or domain knowledge. These tasks model implicit data-label relationships to facilitate representation learning~\cite{medina2022urban}.
Our framework integrates multi-view graph contrastive learning to capture spatial semantics, while dynamically designed prediction and classification tasks model travel traffic dynamics, jointly enhancing road network representation.

\section{Conclusion}\label{sec:con}
This paper proposes \name, a comprehensive road network representation framework that integrates spatial semantics and temporal travel traffic dynamics. \name simultaneously models the functionality of the road network and the movement provided by trajectories. For spatial semantics, we design a mix-hop transition matrix and hyperedges to capture long-range dependencies between roads. 
For temporal travel traffic dynamics, we employ two well-designed tasks to model traffic trends and patterns. 
Extensive experimental results on three real-world datasets across three downstream tasks demonstrate the effectiveness and robustness of \name. 
Furthermore, \name demonstrates satisfactory transferability.
In the future, we will investigate the emergency forecasting problem of road networks in the context of extreme natural disasters, such as typhoons and storms.

\section{Acknowledgments}
This work is supported by the Starry Night Science Fund of Zhejiang University Shanghai Institute for Advanced Study (Grant No. SN-ZJU-SIAS-001), Zhejiang Provincial Natural Science Foundation of China (Grant No. LMS25F020012), the Hangzhou Joint Fund of the Zhejiang Provincial Natural Science Foundation of China under Grant No.LHZSD24F020001, Zhejiang Province High-Level Talents Special Support Program ``Leading Talent of Technological Innovation of Ten-Thousands Talents Program" (No.2022R52046), the Fundamental Research Funds for the Central Universities (No.226-2024-00058), and the advanced computing resources provided by the Supercomputing Center of Hangzhou City University.

\bibliography{aaai2026}

\clearpage

\appendix


\section{A \xspace Preliminary Analysis}
\begin{figure}[tp!]
    \centering
    \includegraphics[width=0.8\columnwidth]{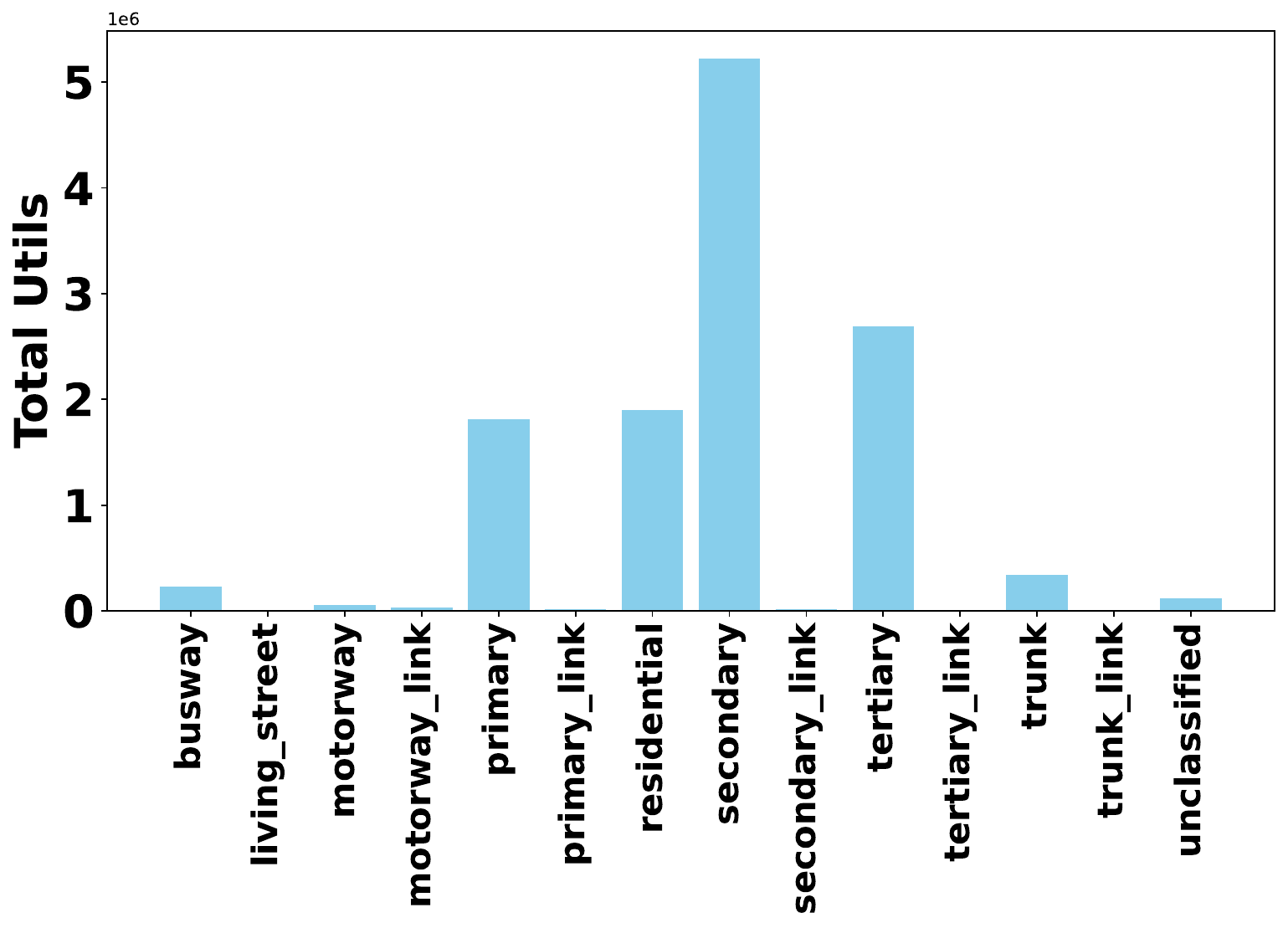}
    \caption{The total volumes of road types within Beijing city.}
    \label{appe:bar}
\end{figure}

Figure~\ref{appe:bar} illustrates the distribution of total access volumes across distinct road types within the Beijing road network, derived from trajectory data. It can be observed that although certain road types exhibit comparable total access volumes, their classifications and functional characteristics differ significantly. We know that access times for individuals vary markedly among road types. Therefore, introducing the dynamically changing volume sequences and modeling them with precision can enhance the temporal dynamic representation of the road network.

\section{B \xspace Details of Methods}

\subsection{B.1 \xspace Details of Pre-processing}
In this section, we introduce the preprocessing pipeline for various input features within the framework of the methods. The first involves node characteristics of the road network. The static features for each road segment are obtained from OpenStreetMap (OSM), including road ID, coordinates, type, length, and the number of lanes. For continuous features, we first discretize them into distinct values, utilizing the discrete values for feature coding. 

The second focuses on edge features of the road network. The angle and Haversine distance between segments connected by edges are computed as edge features. For the angle calculation, we first convert the latitude and longitude coordinates of each road segment into a 3D vector, and then calculate the direction vector and the angle of direction vectors. For any edge $e_{ij}$ linking roads $v_i$ and $v_j$, the Haversine distance is calculated as:
\begin{align}
  a =  \sqrt{ \sin^2 \left( \frac{\Delta\phi}{2} \right) + \cos(\phi_i) \cos(\phi_j) \sin^2 \left( \frac{\Delta\lambda}{2} \right) }, 
\end{align}
\begin{equation}
    d(v_i, v_j) = 2r\arcsin\left( \sqrt{ a } \right) ,
\end{equation}
where $\phi$ denotes the centroid latitude, $\lambda$ is the centroid longitude, and $r$ represents radius of Earth. 

The third addresses the extraction of temporal travel traffic dynamics. The number of visits to each roads across all trajectories is calculated to represent the road traffic volume. This traffic is then categorized into weekdays and weekends over 24 hours based on visit timestamps. This process yields temporal features for two sequence types per road segment.

\section{C \xspace Details of the Experimental Setup}

\subsection{C.1 \xspace Datasets}
All datasets contain two parts of content: road networks and trajectories. The road network data is downloaded from OpenStreetMap\footnote{\url{https://www.openstreetmap.org/}} (OSM), and the trajectory data is collected by vehicles in the corresponding cities. These three datasets can be downloaded from open-source data repositories~\cite{ts_trajgen2023, mao2022jointly}. To make data processing more convenient and efficient, we uniformly convert them into the format defined in the LibCity project~\cite{libcity}. The statistics of the three datasets are shown in Table~\ref{tab:datasets}.

\begin{table}[tp!]
\caption{Statistics of three real-world datasets.}
\label{tab:datasets}
\centering
\small
\begin{tabular}{lccc}
\toprule
Data statistics & Beijing & Porto & San Francisco \\
\midrule
Roads & \num{24355} & \num{10904} & \num{27283} \\
Connections & \num{48171} & \num{25374} & \num{84236} \\
Trajectories & \num{846379} & \num{694602} & \num{233939} \\
Hyperedges & \num{512} & \num{367} & \num{374} \\
\bottomrule
\end{tabular}
\end{table}

\subsection{C.2 \xspace Downstream Tasks} \label{appen: dt}
In this section, we introduce the three downstream tasks for the evaluation. The road speed inference task aims to predict the average speed on roads. We take the learned representations as inputs and train a regression model with only one linear layer. The true value of the average road speed is calculated from the trajectory data.

The trajectory destination prediction task is to predict the end road segment of a trajectory. For trajectories with lengths between 10 and 100, we input the representations of all road segments before the last one into a single-layer LSTM model and use the hidden state of the last time step to predict the destination of the trajectory.

The travel time estimation task is to predict the duration of a trajectory. For trajectories with lengths between 10 and 100, we input the representations of all road segments into a single-layer LSTM model and use the hidden state of the last time step to predict the duration of the trajectory. Specifically, we use seconds as the unit of duration.

\subsection{C.3 \xspace Baselines}
We evaluate \name against several road network representation methods categories. \\
\textbf{\textbullet \ Graph learning methods:}
    (1) Node2Vec~\cite{grover2016node2vec} first employs a random walk algorithm on the graph to obtain sequences of nodes and use the word2vec technique~\cite{mikolov2013efficient} to acquire the representations of graph nodes.
    (2) GCN~\cite{kipf2016semi} is a model that applies the graph convolution computation to road networks. We train it through the graph reconstruction task to obtain the representation of the road network.
    (3) GAE~\cite{velivckovic2017graph} is a model that applies the graph attention mechanism to road networks.  We train it through the graph reconstruction task to obtain the representation of the roads.
    (4) T-GCN~\cite{zhao2019t} integrates the graph convolutional network and the gated recurrent unit to learn the topological structure of the road network and the dynamic changes of road attributes. We utilize it to model the dynamic attributes.

\noindent
\textbf{\textbullet \ Self-supervised RNRL methods:}
    (1) SRN2Vec~\cite{wang2020representation} learns the representation of roads by exploring the relationships among roads that are geographically adjacent and share similar geographical features.
    (2) Toast~\cite{chen2021robust} employs an extended skip-gram model and the context Transformer to capture the travel patterns and travel semantics, and finally unify the two components to get representations.
    (3) JCLRNT~\cite{mao2022jointly}
     employs contrastive learning to conduct a comparative analysis of the interactions between trajectories and the road network, thereby obtaining the representation of the road network.
    (4) TrajRNE~\cite{schestakov2023road} extracts additional road attributes from trajectories and uses trajectories to weight sequence generation algorithm to integrate the contextual information of roads.
    (5) DyToast~\cite{chen2024semantic} further introduces the temporal semantics of trigonometric coding to enhance the integration of temporal dynamics, thus boosting the performance of various time-sensitive downstream tasks.

\subsection{C.4 \xspace Implementation Details} 
\subsubsection{Training Implementation}
DeepHypergraph~\footnote{\url{https://deephypergraph.readthedocs.io/en/latest/index.html}} (DHG) is a deep-learning toolkit based on PyTorch, which applies to the fields of graph neural networks and hypergraph neural networks. It provides a general framework that supports various low-order or high-order information transfer methods, including from vertex to vertex, from vertices in one domain to vertices in another domain, from vertices to hyper-edges, from hyperedges to vertices, and from vertex sets to vertex sets. In this paper, we use the DHG toolkit to complete the work related to hypergraphs.

We use the AdamW~\cite{Loshchilov2017DecoupledWD} optimizer and set the learning rate to 0.001. The number of training epochs for the spatial semantic graph learning is set to 5000, and that of the temporal dynamic learning is set to 100. In addition, we uniformly set the hidden-layer state dimension to 128.

We implement \name with PyTorch 1.3.1. All experiments are run on a 64-bit Linux server with an NVIDIA A6000 GPU. We repeat the experiments 5 times with different random seeds and report the average performance.

\subsubsection{Evaluation Implementation}

We used a single fully connected layer as the classifier and regressor to map the road representations in the road-based tasks. We used the sequence of road representations passed through an LSTM to form the trajectory representation and then predicted the destination and duration with a fully connected layer in the trajectory-based tasks. To compare the generality and validity of the road representations across downstream tasks, we employed simple neural network settings to minimize the influence of other factors.

When evaluating through downstream tasks, we use the five-fold cross-validation for the speed inference task. For the destination prediction task and the travel time estimation task, the trajectory dataset is randomly divided into a 70\% training set and a 30\% test set, with no overlapping trajectories between the two sets. We predict the destination by feeding the representation of roads other than the last one into the model.

\section{D \xspace Additional Contents for Experiments}

\subsection{D.1 \xspace Results on Representation Fusion}
We primarily employ three approaches to fuse representations from different branches: concatenating distinct representations~\cite{schestakov2023road}, directly adding them~\cite{mao2022jointly, NEURIPS2024_15cc8e4a}, and utilizing a gating mechanism~\cite{cao2025holistic}. After conducting a series of comprehensive experimental comparative analyses, we choose the first fusion method. The relevant experimental results have been reported in the main text. Table~\ref{tab:appe-2} and~\ref{tab:appe-3} show the results obtained by adopting the remaining fusion approach. The first approach outperforms in most tasks because it preserves the independence of the original features. Regarding the second and third approaches, their performance shows a significant gap compared with the first approach in the road speed inference task. This may be because the addition of features leads to a reduction in the input dimension, and thus the model fails to fully understand the feature information. In addition, the use of the gating mechanism fusion will bring additional computing resource consumption.

\begin{table}[tp!]
\centering
\tiny
\setlength{\tabcolsep}{1.75mm}
\begin{tabular}{lcccccc}
\toprule
\multirow{2.5}{*}{Datasets} & \multicolumn{2}{c|}{Travel Time Estimation} & \multicolumn{2}{c|}{Speed Inference} & \multicolumn{2}{c}{Destination Prediction}\\
\cmidrule(lr){2-7}
& MAE  ($\mathrel{\downarrow}$) & \multicolumn{1}{c|}{RMSE  ($\mathrel{\downarrow}$)} & MAE  ($\mathrel{\downarrow}$) & \multicolumn{1}{c|}{RMSE  ($\mathrel{\downarrow}$)} & ACC@1  ($\mathrel{\uparrow}$) & \multicolumn{1}{c}{MRR  ($\mathrel{\uparrow}$)} \\
\midrule
Beijing & 240.9307 & \multicolumn{1}{c|}{369.9176} & 3.3846 & \multicolumn{1}{c|}{4.4790} & 0.5573 & 0.6565 \\
Porto & 102.5436 & \multicolumn{1}{c|}{145.9500} & 4.0120 & \multicolumn{1}{c|}{5.3722} & 0.6324 & 0.7642 \\
Xi'an & 236.6982 & \multicolumn{1}{c|}{378.1856} & 4.5397 & \multicolumn{1}{c|}{5.8854} & 0.7602 & 0.8353 \\
\bottomrule
\end{tabular}
\caption{The results of directly adding fusion.}
\label{tab:appe-2}
\end{table}

\begin{table}[tp!]
\centering
\tiny
\setlength{\tabcolsep}{1.75mm}
\begin{tabular}{lcccccc}
\toprule
\multirow{2.5}{*}{Datasets} & \multicolumn{2}{c|}{Travel Time Estimation} & \multicolumn{2}{c|}{Speed Inference} & \multicolumn{2}{c}{Destination Prediction}\\
\cmidrule(lr){2-7}
& MAE  ($\mathrel{\downarrow}$) & \multicolumn{1}{c|}{RMSE  ($\mathrel{\downarrow}$)} & MAE  ($\mathrel{\downarrow}$) & \multicolumn{1}{c|}{RMSE  ($\mathrel{\downarrow}$)} & ACC@1  ($\mathrel{\uparrow}$) & \multicolumn{1}{c}{MRR  ($\mathrel{\uparrow}$)} \\
\midrule
Beijing & 253.4458 & \multicolumn{1}{c|}{386.2368} & 2.7897 & \multicolumn{1}{c|}{3.7643} & 0.5119 & 0.6108 \\
Porto & 107.1870 & \multicolumn{1}{c|}{150.0552} & 3.5959 & \multicolumn{1}{c|}{4.7912} & 0.6359 & 0.7669 \\
Xi'an & 338.6731 & \multicolumn{1}{c|}{456.7167} & 4.5617 & \multicolumn{1}{c|}{5.7872} & 0.7701 & 0.8432
 \\
\bottomrule
\end{tabular}
\caption{The results of the gating mechanism fusion.}
\label{tab:appe-3}
\end{table}

\begin{figure}[tp!]
    \centering
    \includegraphics[width=\linewidth, height=\linewidth]{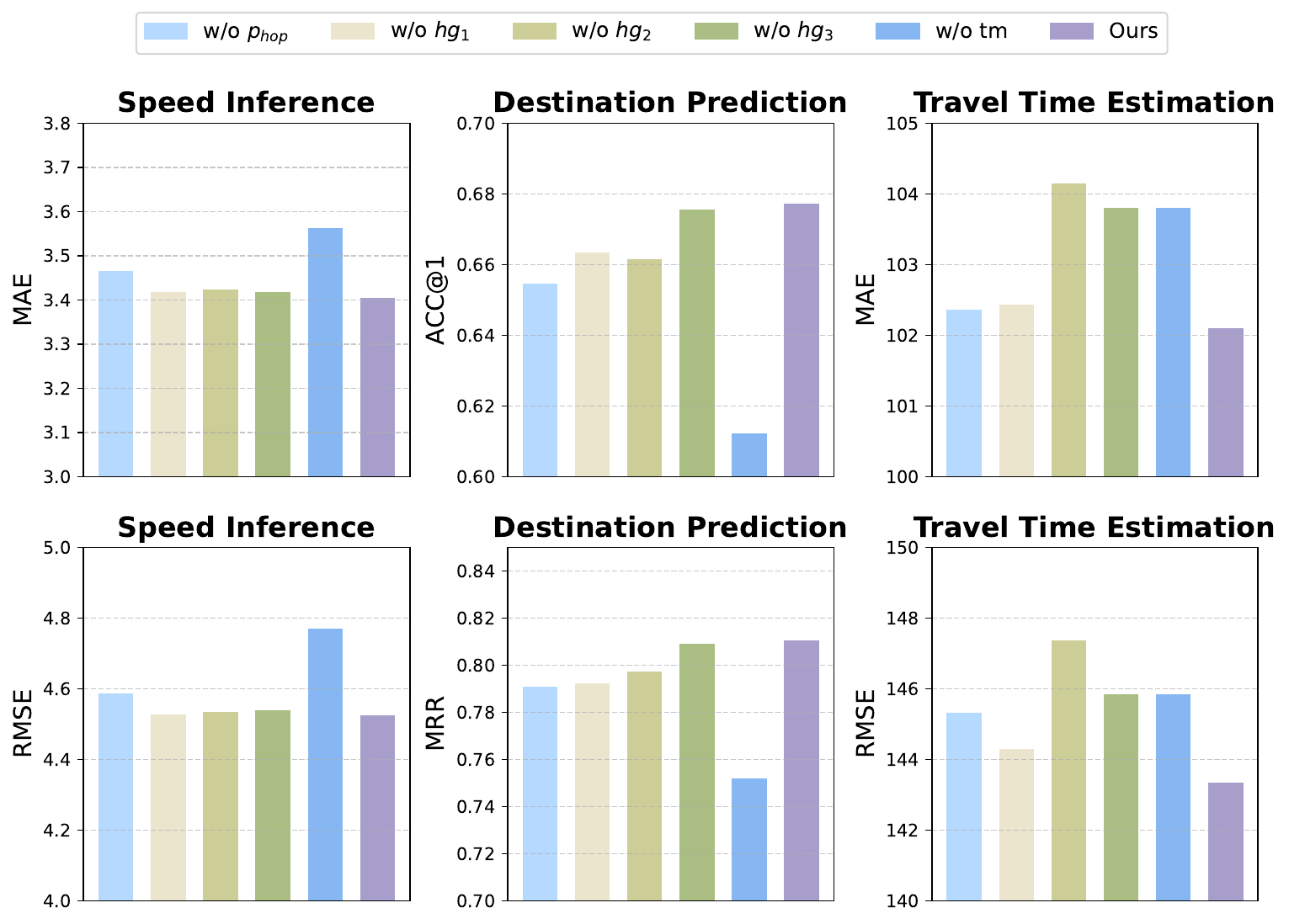}
    \caption{Ablation study on Porto dataset.}
    \label{fig:ab-porto}
\end{figure}

\begin{figure}[tp!]
    \centering
    \includegraphics[width=\linewidth, height=\linewidth]{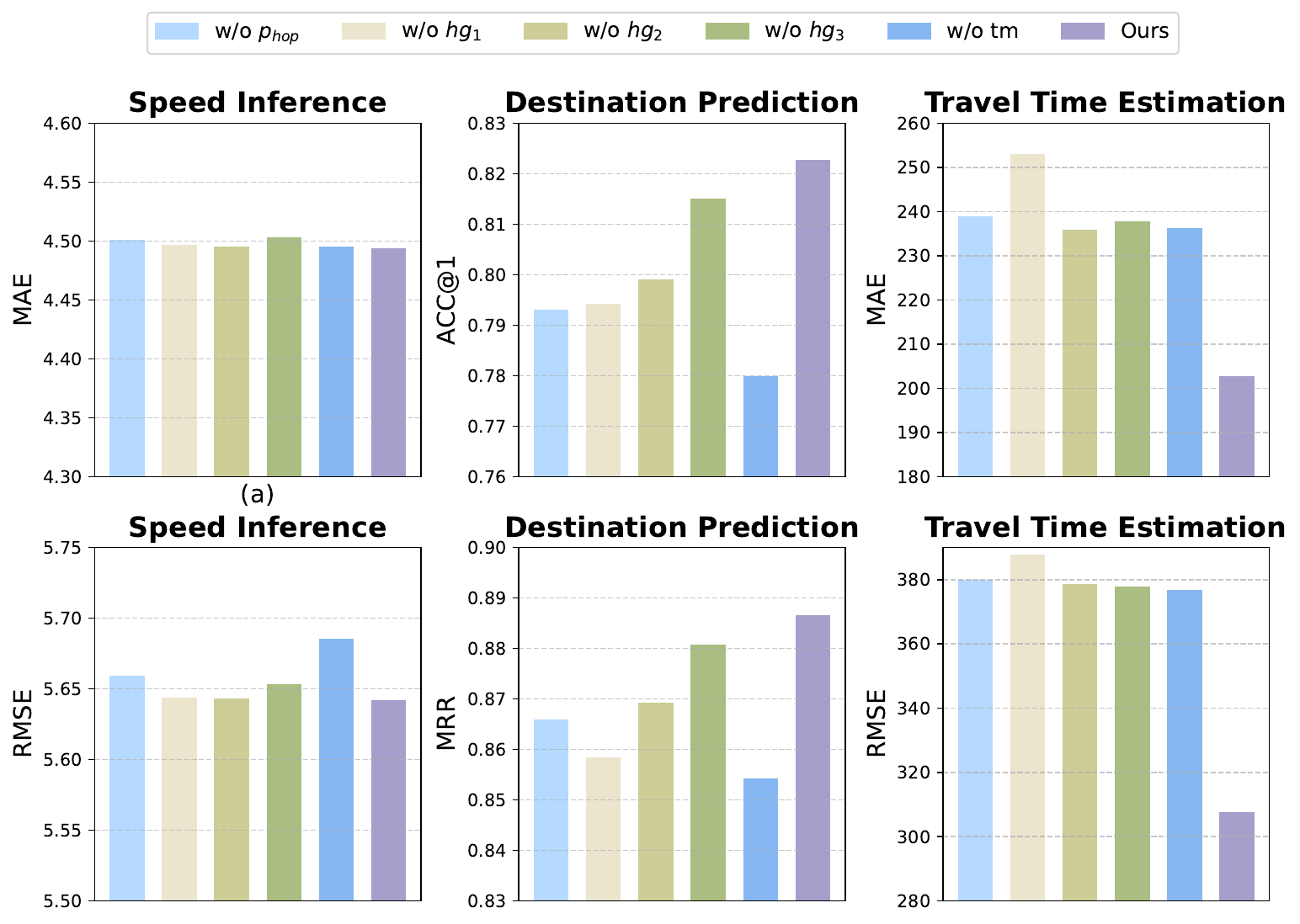}
    \caption{Ablation study on Xi'an dataset.}
    \label{fig:ab-xa}
\end{figure}

\subsection{D.2 \xspace Results on Ablation Studies}

The ablation studies on Porto and Xi'an datasets are shown in Figure~\ref{fig:ab-porto} and~\ref{fig:ab-xa}, respectively. 
Consistent with observations in the Beijing dataset, the removal of all key components brings a performance drop of three downstream tasks. Specifically, the design of the spatial semantics has a more significant impact on road-based tasks, while the design of the temporal travel traffic dynamics integration exerts a greater influence on trajectory-based tasks.

\subsection{D.3 \xspace Efficiency Analysis}
\begin{figure}[tp!]
    \centering
    \includegraphics[width=0.8\linewidth]{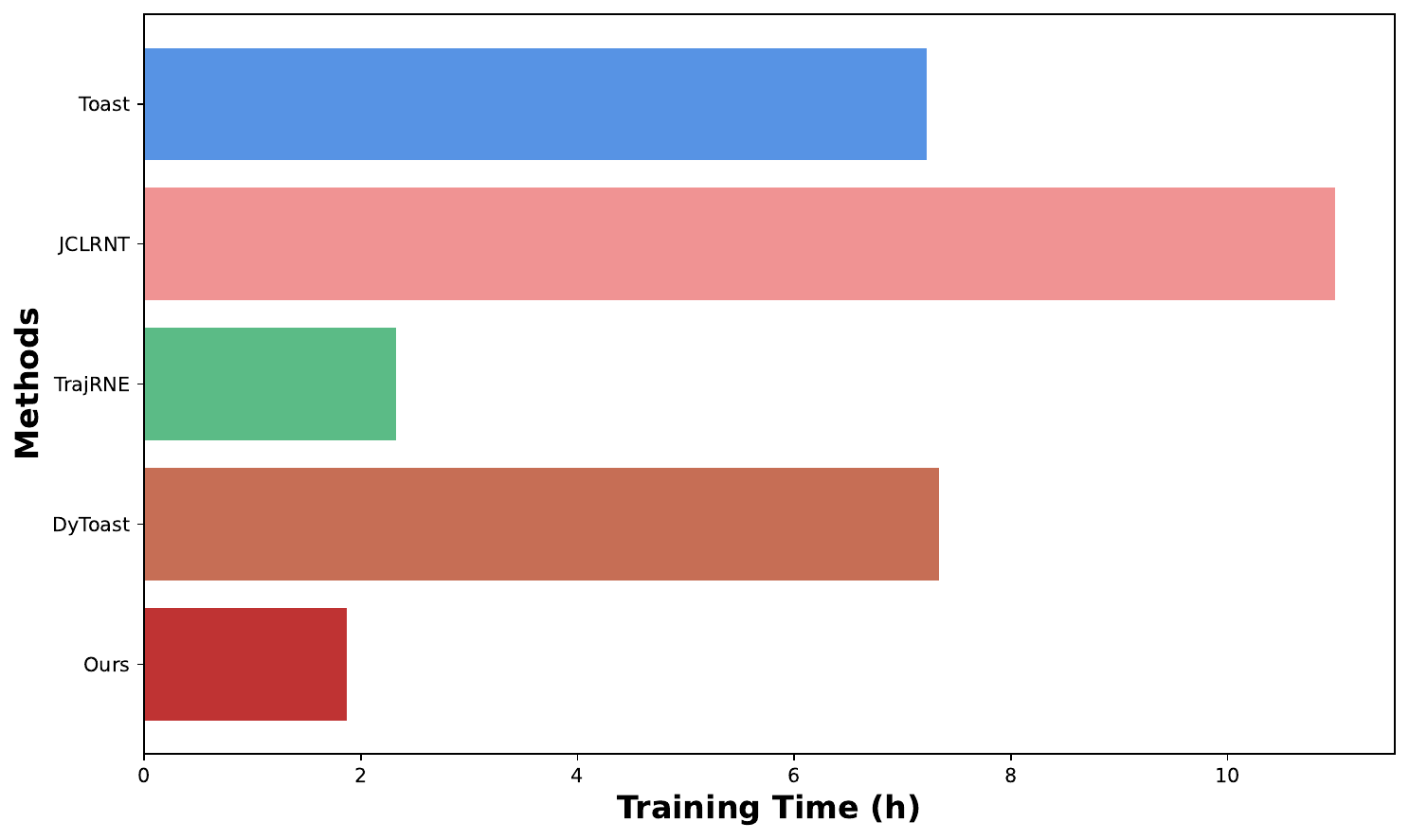}
    \caption{Training Times on Beijing dataset.}
    \label{fig:le}
\end{figure}

We evaluate model efficiency on the Beijing dataset, the largest road network in our study. Figure~\ref{fig:le} compares the training time of \name against other strong baselines, demonstrating the computational advantage of \name. Toast, JCLRNT, and DyToast incur higher learning times due to joint learning trajectory and road network representations. In contrast, TrajRNE and \name leverage trajectory data to optimize road network representation learning. Thus, their training times are much less than others. In conclusion, \name generates high-quality embeddings with substantially reduced training time relative to contemporary techniques.

\begin{table*}[tp!]
\centering
\resizebox{2\columnwidth}{!}{
\begin{tabular}{c|cc|cc|cc|cc|cc|cc|cc|cc|cc}
\toprule
Methods & \multicolumn{2}{c|}{Node2Vec} & \multicolumn{2}{c|}{GCN} & \multicolumn{2}{c|}{GAE} & \multicolumn{2}{c|}{T-GCN} & \multicolumn{2}{c|}{SRN2Vec} & \multicolumn{2}{c|}{Toast} & \multicolumn{2}{c|}{JCLRNT} & \multicolumn{2}{c|}{TrajRNE} & \multicolumn{2}{c}{DyToast} \\
 \cmidrule(r){1 - 1} \cmidrule(r){2 - 3} \cmidrule(r){4 - 5} \cmidrule(r){6 - 7} \cmidrule(r){8 - 9} \cmidrule(r){10 - 11} \cmidrule(r){12 - 13} \cmidrule(r){14 - 15} \cmidrule(r){16 - 17} \cmidrule(r){18 - 19}   
 Metrics & MAE &  RMSE & MAE & RMSE & MAE & RMSE & MAE & RMSE & MAE & RMSE & MAE & RMSE & MAE & RMSE & MAE & RMSE & MAE & RMSE \\
\midrule
Beijing & 0.00$^*$ & 0.00$^*$ & 0.00$^*$ & 0.00$^*$ & 0.00$^*$ & 0.00$^*$ & 0.00$^*$ & 0.00$^*$ & 0.00$^*$ & 0.00$^*$ & 0.00$^*$ & 0.00$^*$ & 0.00$^*$ & 0.00$^*$ & 0.00$^*$ & 0.00$^*$ & 0.00$^*$ & 0.00$^*$ \\
\midrule
Porto & 0.00$^*$ & 0.00$^*$ & 0.00$^*$ & 0.00$^*$ & 0.00$^*$ & 0.00$^*$ & 0.00$^*$ & 0.00$^*$ & 0.00$^*$ & 0.00$^*$ & 0.00$^*$ & 0.00$^*$ & 0.00$^*$ & 0.00$^*$ & 0.00$^*$ & 0.00$^*$ & 0.00$^*$ & 0.00$^*$ \\
\midrule
Xi'an & 0.00$^*$ & 0.00$^*$ & 0.00$^*$ & 0.00$^*$ & 0.00$^*$ & 0.00$^*$ & 0.00$^*$ & 0.00$^*$ & 0.00$^*$ & 0.00$^*$ & 0.00$^*$ & 0.00$^*$ & 0.11 & 0.00$^*$ & 0.00$^*$ & 0.00$^*$ & 0.00$^*$ & 0.00$^*$ \\
\bottomrule
\end{tabular}
}
\caption{Significance Analysis on Speed Inference. $^*$ indicates $p \leq 0.01$.}
\label{tab:speed_inference_pvalues_2dp}
\end{table*}

\begin{table*}[tp!]
\centering
\resizebox{2\columnwidth}{!}{
\begin{tabular}{c|cc|cc|cc|cc|cc|cc|cc|cc|cc}
\toprule
Methods & \multicolumn{2}{c|}{Node2Vec} & \multicolumn{2}{c|}{GCN} & \multicolumn{2}{c|}{GAE} & \multicolumn{2}{c|}{T-GCN} & \multicolumn{2}{c|}{SRN2Vec} & \multicolumn{2}{c|}{Toast} & \multicolumn{2}{c|}{JCLRNT} & \multicolumn{2}{c|}{TrajRNE} & \multicolumn{2}{c}{DyToast} \\
 \cmidrule(r){1 - 1} \cmidrule(r){2 - 3} \cmidrule(r){4 - 5} \cmidrule(r){6 - 7} \cmidrule(r){8 - 9} \cmidrule(r){10 - 11} \cmidrule(r){12 - 13} \cmidrule(r){14 - 15} \cmidrule(r){16 - 17} \cmidrule(r){18 - 19}   
 Metrics & MAE &  RMSE & MAE & RMSE & MAE & RMSE & MAE & RMSE & MAE & RMSE & MAE & RMSE & MAE & RMSE & MAE & RMSE & MAE & RMSE \\
\midrule
Beijing & 0.01$^*$ & 0.01$^*$ & 0.00$^*$ & 0.00$^*$ & 0.00$^*$ & 0.00$^*$ & 0.00$^*$ & 0.00$^*$ & 0.04 & 0.00$^*$ & 0.01$^*$ & 0.00$^*$ & 0.01$^*$ & 0.00$^*$ & 0.39 & 0.53 & 0.08 & 0.27 \\
\midrule
Porto & 0.01$^*$ & 0.00$^*$ & 0.00$^*$ & 0.00$^*$ & 0.00$^*$ & 0.00$^*$ & 0.00$^*$ & 0.00$^*$ & 0.30 & 0.31 & 0.30 & 0.01$^*$ & 0.39 & 0.02 & 0.18 & 0.03 & 0.33 & 0.15 \\
\midrule
Xi'an & 0.00$^*$ & 0.00$^*$ & 0.00$^*$ & 0.00$^*$ & 0.00$^*$ & 0.00$^*$ & 0.00$^*$ & 0.00$^*$ & 0.00$^*$ & 0.00$^*$ & 0.00$^*$ & 0.00$^*$ & 0.00$^*$ & 0.00$^*$ & 0.17 & 0.13 & 0.00$^*$ & 0.00$^*$ \\
\bottomrule
\end{tabular}
}
\caption{Significance Analysis on Travel Time Estimation. $^*$ indicates $p \leq 0.01$.}
\label{tab:travel_time_significance}
\end{table*}

\begin{table*}[tp!]
\centering
\resizebox{2\columnwidth}{!}{
\begin{tabular}{c|cc|cc|cc|cc|cc|cc|cc|cc|cc}
\toprule
Methods & \multicolumn{2}{c|}{Node2Vec} & \multicolumn{2}{c|}{GCN} & \multicolumn{2}{c|}{GAE} & \multicolumn{2}{c|}{T-GCN} & \multicolumn{2}{c|}{SRN2Vec} & \multicolumn{2}{c|}{Toast} & \multicolumn{2}{c|}{JCLRNT} & \multicolumn{2}{c|}{TrajRNE} & \multicolumn{2}{c}{DyToast} \\
 \cmidrule(r){1 - 1} \cmidrule(r){2 - 3} \cmidrule(r){4 - 5} \cmidrule(r){6 - 7} \cmidrule(r){8 - 9} \cmidrule(r){10 - 11} \cmidrule(r){12 - 13} \cmidrule(r){14 - 15} \cmidrule(r){16 - 17} \cmidrule(r){18 - 19}   
 Metrics & ACC@1 &  MRR & ACC@1 & MRR & ACC@1 & MRR & ACC@1 & MRR & ACC@1 & MRR & ACC@1 & MRR & ACC@1 & MRR & ACC@1 & MRR & ACC@1 & MRR \\
\midrule
Beijing & 0.00$^*$ & 0.00$^*$ & 0.00$^*$ & 0.00$^*$ & 0.00$^*$ & 0.00$^*$ & 0.00$^*$ & 0.00$^*$ & 0.00$^*$ & 0.00$^*$ & 0.00$^*$ & 0.00$^*$ & 0.00$^*$ & 0.00$^*$ & 0.00$^*$ & 0.00$^*$ & 0.00$^*$ & 0.00$^*$ \\
\midrule
Porto & 0.00$^*$ & 0.00$^*$ & 0.00$^*$ & 0.00$^*$ & 0.00$^*$ & 0.00$^*$ & 0.00$^*$ & 0.00$^*$ & 0.01$^*$ & 0.00$^*$ & 0.01$^*$ & 0.01$^*$ & 0.00$^*$ & 0.00$^*$ & 0.03 & 0.03 & 0.00$^*$ & 0.00$^*$ \\
\midrule
Xi'an & 0.00$^*$ & 0.00$^*$ & 0.00$^*$ & 0.00$^*$ & 0.00$^*$ & 0.00$^*$ & 0.00$^*$ & 0.00$^*$ & 0.01$^*$ & 0.00$^*$ & 0.04 & 0.03 & 0.00$^*$ & 0.00$^*$ & 0.03 & 0.00$^*$ & 0.02 & 0.01$^*$ \\
\bottomrule
\end{tabular}
}
\caption{Significance Analysis on Destination Prediction. $^*$ indicates $p \leq 0.01$.}
\label{tab:destination_prediction_significance}
\end{table*}

\begin{figure}[bp!]
    \centering
    \includegraphics[width=\linewidth, height=0.7\linewidth]{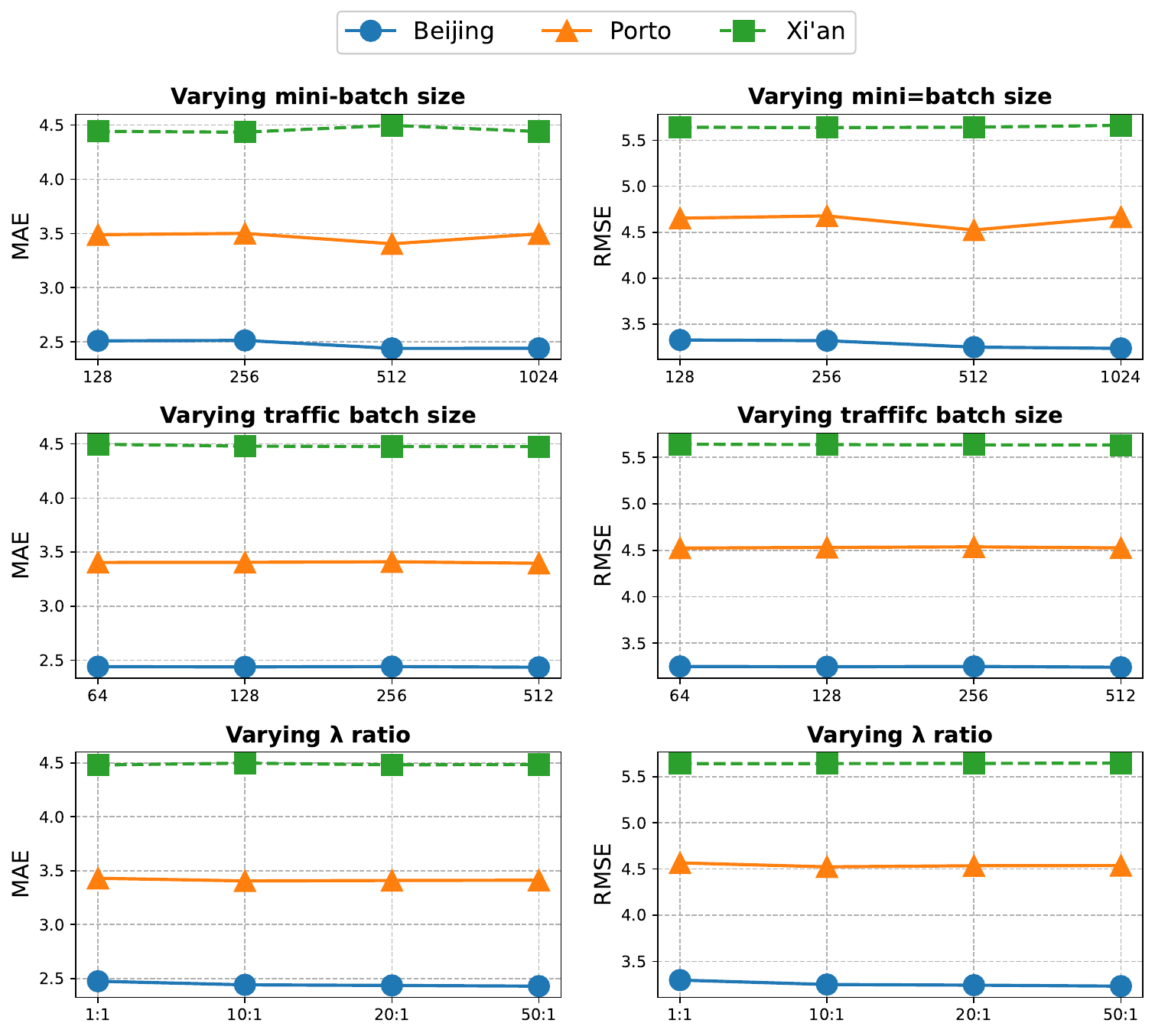}
    \caption{Parameter sensitivity on speed inference.}
    \label{fig:pa-si}
\end{figure}

\begin{figure}[bp!]
    \centering
    \includegraphics[width=\linewidth, height=0.75\linewidth]{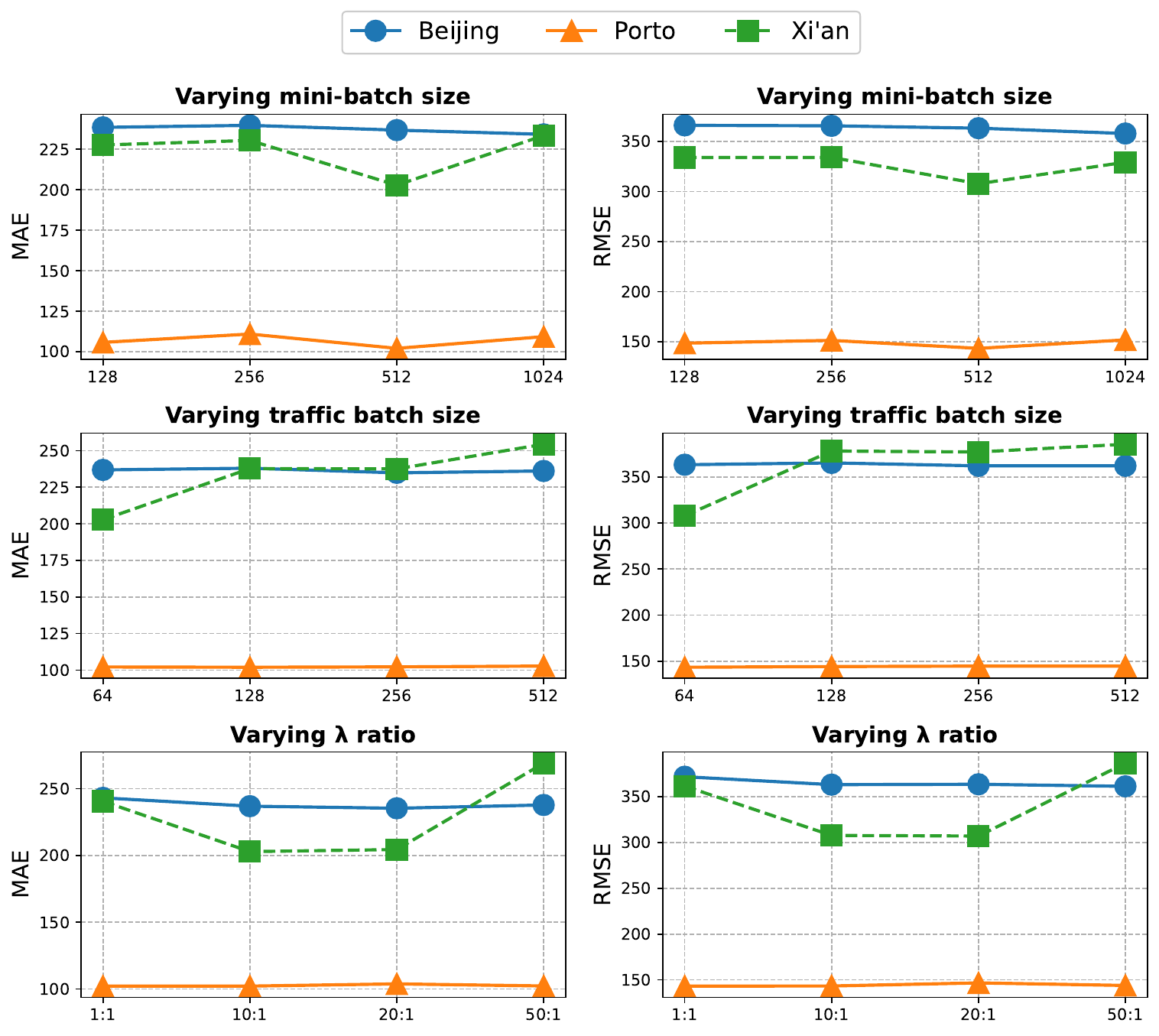}
    \caption{Parameter sensitivity on travel time estimation.}
    \label{fig:pa-te}
\end{figure}

\subsection{D.4 \xspace Significance Analysis}
We conduct T-tests to assess performance improvement in \name. Tables~\ref{tab:speed_inference_pvalues_2dp}, ~\ref{tab:travel_time_significance}, and ~\ref{tab:destination_prediction_significance} present significance test results for the three tasks. Statistical analyses indicate significantly enhanced performance of \name relative to most comparative methods.

\subsection{D.5 \xspace Results on parameters varying}

Figure~\ref{fig:pa-si} and ~\ref{fig:pa-te} show the results of varying parameters on speed inference and travel time estimation tasks. We can find a consistent conclusion with the main text described.

\section{E \xspace Limitations}
Our largest dataset contains 25,000 nodes, which is 3\textasciitilde4 times larger than most benchmarks. However, the larger-scale cities (such as those with 100,000 nodes) may still trigger out-of-memory errors in our proposed method. In addition, while \name has strong transferability, some modifications (e.g., removing some modules) inevitably compromise performance. 

\end{document}